\documentclass{article}
\pdfoutput=1

\PassOptionsToPackage{numbers, compress}{natbib}
\usepackage[final]{neurips_2022}

\usepackage[utf8]{inputenc} % allow utf-8 input
\usepackage[T1]{fontenc}    % use 8-bit T1 fonts
\usepackage{hyperref}       % hyperlinks
\usepackage{url}            % simple URL typesetting
\usepackage{booktabs}       % professional-quality tables
\usepackage{amsfonts}       % blackboard math symbols
\usepackage{nicefrac}       % compact symbols for 1/2, etc.
\usepackage{microtype}      % microtypography
\usepackage[table]{xcolor}         % colors

\usepackage{graphicx}
\usepackage{amsmath}
\usepackage{amssymb}
\usepackage{caption}
\usepackage{subcaption}

\usepackage{algorithm}
\usepackage[noend]{algpseudocode}
\usepackage{mathtools}
\usepackage{amsthm}
\usepackage{bm}
\usepackage{wrapfig}

\DeclareMathOperator*{\argmax}{arg\,max}
\DeclareMathOperator*{\argmin}{arg\,min}
\newcommand\norm[1]{\left \Vert #1 \right \Vert}
\newcommand\braces[1]{\left \lbrace #1 \right \rbrace}

\newcommand\func[2]{#1 \left( #2 \right)}

\usepackage{pifont}

\usepackage{multirow}
\usepackage{rotating}

\newcommand{\regularizer}[1]{$\mathcal{L}_{\mathcal{E}}^#1$}
\newcommand{\regularizerhat}[1]{$\mathcal{L}_{\mathcal{E}}^{\hat{#1}}$}
\newcommand{\task}{$\mathcal{L}_{\text{task}}$}
\newcommand*\samethanks[1][\value{footnote}]{\footnotemark[#1]}

\usepackage{enumitem}

\newtheorem{theorem}{Theorem}[section]

\newtheorem{lemma}[theorem]{Lemma}

\usepackage{lmodern}

\title{UDC: \underline{U}nified \underline{D}NAS for \underline{C}ompressible TinyML Models for Neural Processing Units}

\author{%
  Igor Fedorov\thanks{Work done while I.F., R. M., H. T., and M. M. were at Arm Inc.} \\
  Meta AI \\
  Menlo Park, CA 94025 \\
  \texttt{fedorov.uofi@gmail.com} \\
   \And
   Ramon Matas\samethanks\\
   NVIDIA \\
   Santa Clara, CA 95050 \\
   \texttt{ramonm@nvidia.com} \\
  \And
   Hokchhay Tann\samethanks \\
   Tenstorrent \\
   Boston, MA 01721 \\
   \texttt{ctann@tenstorrent.com} \\
   \And
   Chuteng Zhou \\
   Arm Inc.\\
   Boston, MA 01721 \\
   \texttt{chu.zhou@arm.com} \\
   \And
   Matthew Mattina\samethanks \\
   Tenstorrent \\
   Boston, MA 01721 \\
   \texttt{mmattina@tenstorrent.com} \\
   \And
   Paul N. Whatmough\samethanks \\
   Qualcomm AI Research\\
   Cambridge, MA 02140 \\
   \texttt{pwhatmou@qti.qualcomm.com} \\
}

\begin{document}

\maketitle

\begin{abstract}
Deploying TinyML models on low-cost IoT hardware is very challenging, due to limited device memory capacity. Neural processing unit (NPU) hardware address the memory challenge by using model compression to exploit weight quantization and sparsity to fit more parameters in the same footprint. However, designing compressible neural networks (NNs) is challenging, as it expands the design space across which we must make balanced trade-offs. This paper demonstrates \underline{U}nified \underline{D}NAS for \underline{C}ompressible (\texttt{UDC}) NNs, which explores a large search space to generate state-of-the-art \textit{compressible} NNs for NPU. 
ImageNet results show \texttt{UDC} networks are up to $3.35\times$ smaller (iso-accuracy) or $6.25\%$ more accurate (iso-model size) than previous work.
\end{abstract}
\section{Introduction}

IoT applications demand \textit{TinyML} models that fit on highly-constrained hardware (HW), with limited memory and compute power~\citep{banbury2021micronets,fedorov2019sparse,gupta2017protonn,kumar2017resource}. Canonical TinyML tasks include visual wakewords, audio keywords, anomaly detection, speech enhancement, and image classification~\cite{banbury2021micronets,banbury2021mlperf,fedorov2020tinylstms,visual_wake,lin2020mcunet}. 
Traditional microcontroller units (MCUs) are not well suited to meet the memory and compute challenges of TinyML, so HW vendors offer specialized processors for neural network (NN) inference, called neural processing units (NPUs)~\cite{ethos-u55,alif}. 

%Software (SW) inference runtimes for
%When deploying NNs on 
%conventional 
MCU inference runtimes~\cite{tflitemicro,utensor}
%the software inference runtime 
%only support 8-bit quantization, with 
do not implement model compression, which is slow in software.
Hence MCUs do not benefit from sub 8-bit quantization or 
%Thus, there is no benefit 
%does not reduce the deployed model size. 
%Similarly, 
unstructured pruning \cite{han2015learning}.
%on current MCU runtimes~\cite{tflitemicro,utensor}.
%has no benefit on MCUs, as there is neither HW nor software support for compression or sparse math
%However, NPUs~\cite{ethos-u55,alif} 
%%Model compression is an important feature of NPUs like 
%%(e.g. ARM Ethos U55
%support HW model compression, which allows NN weights to be stored in a reduced memory footprint and decompressed by the NPU at inference time. 
%Model compression offers an enormous advantage for TinyML tasks, where meager Flash memory is usually the bottleneck to increasing accuracy~\cite{fedorov2019sparse,lin2020mcunet,gupta2017protonn,kumar2017resource}. 
%A smaller memory footprint also reduces memory access energy---critical for battery powered IoT devices \cite{li2019chip,ahmad2020superslash,han2015learning,practical2020tinyml}. 
In contrast, NPUs with HW model compression~\cite{ethos-u55,alif} do benefit from both optimizations
%sub-byte quantization and unstructured pruning 
\cite{ethos-vela}. 
For example, the Arm Ethos-U55 NPU compiler (Vela~\cite{ethos-vela}) encodes weight tensors using two components: 1) a binary mask marking non-zero elements (run-length and Golomb-Rice compressed), and 2) the non-zero values (Golomb-Rice compressed).
A model aggressively pruned to have many zeros and quantized to low bitwidths can be deployed in very little device memory. 
At inference time, weights are then decompressed by the NPU on demand.
%during inference, so 
%Both tuple components are then further compressed, using run-length for the binary mask, and Golomb-Rice coding for the non-zeros.

\begin{figure*}
\centering
%\hspace{-0.5em}
\includegraphics[width=\linewidth]{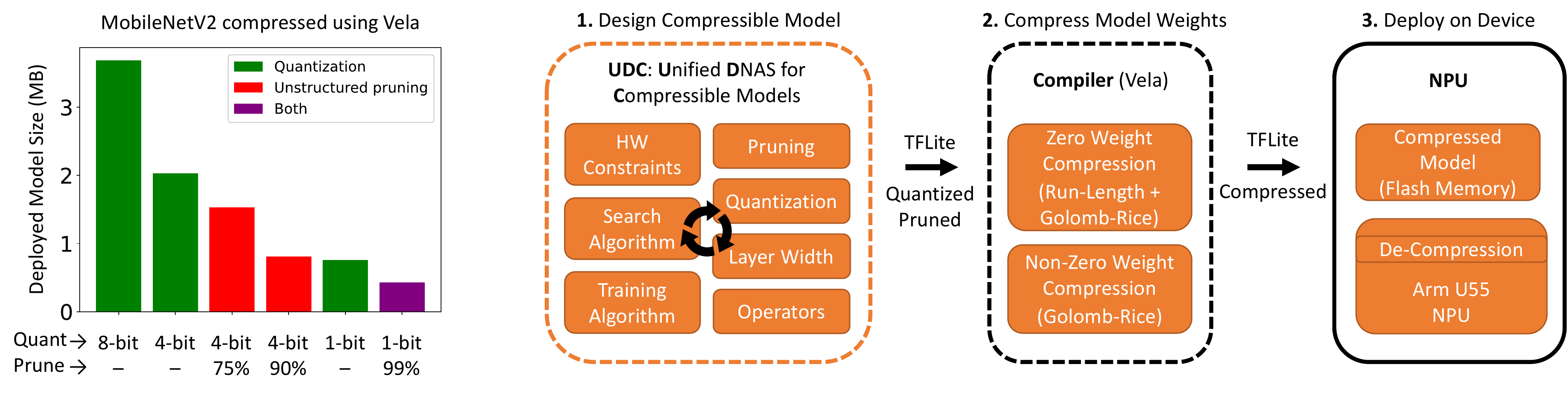}
% \vspace{-1em}
\caption{\textbf{Left:} NPU weight compression allows significant deployed model size reduction for TinyML.
Compressed MBNetV2 size scales with increasingly aggressive quantization and pruning choices.
\textbf{Right:} \texttt{UDC} designs compressible models, tailored to NPU deployment.}
\label{fig:udc overview}
\end{figure*}

NPU compression is an enormous advantage for TinyML, where meager Flash memory limits model size and therefore accuracy~\cite{fedorov2019sparse,lin2020mcunet,gupta2017protonn,kumar2017resource}. 
A smaller memory footprint also reduces memory access energy, critical for battery powered IoT devices \cite{li2019chip,ahmad2020superslash,han2015learning,practical2020tinyml}.  
Fig. \ref{fig:udc overview} shows how model size scales with quantization, pruning, and both, when compressing a MBNetV2 \cite{howard2017mobilenets} for deployment using Vela, down to a $9.3$$\times$ reduction in model size for 1-bit quantization and 1\% non-zero weights in the extreme.

%% Whats wrong with dnas for tinyml??
To fully exploit NPU model compression, we define our problem statement as follows:%we state the following problem statement:
\begin{align}\tag{P1}\label{P1}\small
    % \textbf{DNAS generated NNs should consume the minimum storage space.}
    \textbf{Automatically design compressible NNs with the smallest model size.}
\end{align}\normalsize
%The steps 
%Practitioners typically design deployable models in two steps.
%take to 
Typically, the model architecture itself is developed first,
%state-of-the-art deployable NNs either 
%are: 1) Given a task-specific dataset, design an NN architecture which performs well on that task 
either through manual trial-and-error or by neural architecture search (NAS) \cite{liberis2021munas, dong2019network, ye2018rethinking, he2017channel,molchanov2016pruning,yang2017designing,fedorov2020tinylstms,wan2020fbnetv2,wu2019fbnet}.
%, 2) Given the NN architecture, 
Then, model conditioning techniques, e.g., quantization~\cite{Uhlich2020Mixed,yu2020search,Wang_2020_CVPR,yang2020automatic,wang2018haq} and unstructured pruning
\cite{choi2020learning,kusupati2020soft,han2015deep,han2015learning,Lin2020Dynamic}, are applied before 
%3) 
deploying to the target HW. 
%For NPUs, 
%integer quantization is an essential step
%There are two additional considerations for NPU. 
Furthermore, the generated model must also be small enough to deploy on the NPU device;
merely regularizing size does not guarantee that the models found will fit into device Flash memory.
It is essential that:
\begin{align}\tag{P2}\label{P2}\small
    % \textbf{DNAS generated NNs must not exceed a hard constraint on model size.}
    \textbf{Generated NNs do not exceed a hard constraint on (compressed) model size.}
\end{align}
%the memory on the device. 

This paper describes \texttt{UDC} (Fig. \ref{fig:udc overview}),
%shows our UDC framework 
which merges the model design and conditioning steps by 
%1) and 2) and 
conducting a joint search over NN architecture, weight bitwidths, and sparsity rates. 
\texttt{UDC} builds upon differentiable NAS (DNAS), which exploits efficient weight sharing to solve \eqref{P1}, while addressing key challenges like how to explore the design space while still addressing \eqref{P2} \cite{liu2018darts,pham2018efficient}. 
The search space is HW-compression-aware,
%considering
%because 
including only model conditioning techniques supported by
%which are compatible with 
the real-world Vela NPU compiler.
%are considered 
We explicitly exclude low-rank matrix factorization~\cite{idelbayev2020low} and 
%and therefore 
non-uniform floating-point quantization~\cite{yang2020automatic,wang2018haq}, which are not supported. 
%NPUs for IoT only support integer convolutions, so
%The following summarizes our contributions.
The contributions of this paper are further summarized below.

%\textbf{Unstructured Pruning \& Quantization in DNAS} 
\textbf{Joint network architecture and conditioning search}
We extend the DNAS formalism to learn layer-wise weight sparsity levels. We present a method for searching for sparsity levels in conjunction with layer bitwidths. We show how to maximize weight sharing while jointly searching over sparsity, bitwidth, and layer width, as well as provide a differentiable and easily computable measure of compressibility for the DNAS algorithm to optimize.

\textbf{Novel search algorithm} 
Previous 
%DNAS 
work (e.g. \cite{dong2019searching,dong2019network}) fails to effectively trade-off accuracy with model size \eqref{P2} in our search space (Table \ref{table:ablation}, first row). 
%More specifically, 
\texttt{UDC} addresses this, with the following improvements:
%overcomes three related issues,
%with previous DNAS work, 
%\textit{viz.}
%and propose a novel algorithm which resolves them, namely 
1) guarantees that the search yields a model which satisfies specified HW constraints (Sec. \ref{section:dnas}), 2) provides control over exploration-exploitation 
%characteristics 
(Sec. \ref{subsection:exploration-exploitation}), and 3) avoids over-regularization from
%due to 
biased Gumbel-softmax approximation (Sec. \ref{section:dnas}, \ref{subsection:rejection sampling}).

\textbf{Novel sparse, low-bitwidth representation and training algorithm} 
\texttt{UDC} yields compressible models that are difficult to train. 
We identify the root cause of the problem and propose a solution using a novel weight representation (Sec. \ref{sec:training sparse quantized models}).

%\textbf{Extremely Small, SOTA NPU-Deployable TinyML Models} 
\textbf{SOTA NPU-deployable TinyML models} 
Using \texttt{UDC}, we 
%find compact, 
demonstrate compressible NNs Pareto dominant over
%with respect to the 
prior work on CIFAR100, ImageNet, and 
%div2k/set5/set14 
super resolution (SR) tasks (Fig. \ref{figu:imagenet and cifar100 and super resolution}) \cite{krizhevsky2009learning,ILSVRC15,agustsson2017ntire,Huang-CVPR-2015}. 
%We validate the reported 
Compression gains are validated by deploying to Ethos-U55 NPU.
\newpage
\section{Related work}
%\vspace{-1em}
\begin{wraptable}{r}{0.55\textwidth}
% \vspace{-0.8em}
\caption{\texttt{UDC} features vs. related work.}
\label{table:udc features}
\centering
\resizebox{0.55\textwidth}{!}{%
\begin{tabular}{lccccccccc} 
\toprule
& \rotatebox{90}{\textbf{MCUnet} \cite{lin2020mcunet}} & \rotatebox{90}{\textbf{APQ} \cite{Wang_2020_CVPR}} & \rotatebox{90}{\textbf{HAQ} \cite{wang2018haq}} & \rotatebox{90}{\textbf{Yang et al.} \cite{yang2020automatic}} & \rotatebox{90}{\textbf{Gong et al.} \cite{gong2019mixed}} & \rotatebox{90}{\textbf{Choi et al.} \cite{choi2020learning}} & \rotatebox{90}{\textbf{Uhlich et al.} \cite{Uhlich2020Mixed}} & \rotatebox{90}{\textbf{FBNetV2} \cite{wan2020fbnetv2}} & \rotatebox{90}{\textbf{\textcolor{red}{\textbf{\texttt{UDC} (Ours)}}}} \\
\midrule
Width/Operator/Depth & \checkmark & \checkmark & & & &   & & \checkmark & \checkmark  \\
Bitwidth/Sparsity & & &  & \checkmark & & \checkmark & & & \checkmark\\
HW Constraint Guarantee & \checkmark & \checkmark & & \checkmark & \checkmark & & \checkmark & & \checkmark \\
Deployable w/Integer Math & \checkmark & \checkmark & & & \checkmark & \checkmark & \checkmark & \checkmark & \checkmark \\
TinyML Size ($<1.25$MB) & \checkmark & & \checkmark & \checkmark & \checkmark & \checkmark & \checkmark & & \checkmark \\
\bottomrule
\end{tabular}
}
% \vspace{-1em}
\end{wraptable}
Table \ref{table:udc features} contrasts \texttt{UDC} with 
%the most 
relevant previous work. 
MCUnet \cite{lin2020mcunet} is the closest to \texttt{UDC}, 
%in terms of designing accurate NNs for large scale datasets like ImageNet while 
as it targets the TinyML HW form-factor with severely limited memory and includes results on large scale datasets like ImageNet.
However, \texttt{UDC} is fundamentally different to MCUNet: 1) unlike MCUnet, \texttt{UDC} also performs per-layer 
%a search of 
bitwidths and sparsity rate search to exploit HW compression, 
2) \texttt{UDC} is a DNAS algorithm whereas MCUnet uses evolutionary search of a pre-trained once-for-all (OFA) supernet \cite{cai2019once}. 
In particular, the \texttt{UDC} search space targets compressible models and is much more diverse than that of MCUNet; training a single OFA supernet for our search space would be infeasible (Sec. \ref{section:ablations}).

% PNW TODO

\citet{yang2020automatic} also demonstrate learning of per-layer bitwidth and sparsity rate, but otherwise differ significantly
%from \texttt{UDC} 
in that they: 1) do not include storage of the (required) pruning mask in the reported model size, which can even dominate
%often a significant portion of the model size 
at high sparsity rates \cite{MLSYS2020_d2ddea18};
%and would inflate the reported model size 
%for practical deployment; 
2) only consider model conditioning, without accompanying NN width, depth, or operator search;
3) employ floating-point non-uniform quantization, not deployable on TinyML MCU and NPU platforms which only support integer operations. 

APQ proposes a multi-stage algorithm to search over layer width and bitwidth \cite{Wang_2020_CVPR}. Compared to APQ, \texttt{UDC}: 1) optimizes over sparsity rates whereas APQ does not, and 2) produces models $3$--$9\times$ smaller than APQ. 
Other relevant works include \citet{gong2019mixed}, which searches over width and bitwidth but not sparsity, \citet{choi2020learning}, which searches over bitwidth and sparsity but not layer widths and without addressing \eqref{P2}, and \citet{Uhlich2020Mixed} which searches over only bitwidths.
\section{Modelling design decisions}
\label{sec:searching from compressible models}
%\vspace{-1em}

We use bold-face to denote vectors/tensors, and $z[k]$ to mean the $k$'th element of vector $\mathbf{z}$. For a layer with input $\bm{x}$ and parameters $\bm{\theta}$, we denote its output by $\func{f}{\bm{x},\bm{\theta}}$. We say $\bm{z} \in \mathbb{R}^K$ follows a categorical distribution parameterized by $\bm{\pi}$,  i.e. $\bm{z} \sim \func{\text{Cat}}{\bm{\pi}}$, if $p(z[k] = 1) = \pi[k]$. 

\textbf{Width selection}
%Different 
Layer widths are modelled as $f(\bm{x},\bm{\theta}) \odot \bm{w}, \norm{\bm{w}}_0 / \vert \bm{w} \vert = \rho$, where $\bm{w}$ is a binary mask to toggle any given channel, $\rho$ is the fraction of non-zero channels, $\odot$ is element-wise multiplication, $\norm{\bm{w}}_0$ is the $\ell_0$ pseudo-norm which counts the number of non-zeros in $\bm{w}$, and $\vert \bm{w} \vert$ is the number of elements in $\bm{w}$. We adopt the convention of setting the first $\rho$ fraction of the channels of $\bm{w}$ to $1$ \cite{wan2020fbnetv2}. Different choices $\braces{\rho_1,\cdots,\rho_{K_w}}$ of layer width can be modelled by the random variable (RV) 
\begin{align}
\bm{w}(\bm{\pi_w}) = \sum_{k=1}^{K_w} z_w[k] \bm{w_k}, \bm{z_w} \sim \func{\text{Cat}}{\bm{\pi_w}}, \norm{\bm{w_k}}_0 / \vert \bm{w_k} \vert = \rho_k. 
\end{align}
We refer to $\bm{z_w}$ as a decision variable and to $\rho_k$ as an option. 

% \vspace{-2em}
\paragraph{Sparsity} Sparse weight tensors are expressed as $\bm{\theta} \odot \bm{m}, \norm{\bm{m}}_0 / \vert \bm{m} \vert = s$
where $\bm{m}$ is a binary mask and $s$ is the fraction of non-zeros. We set the non-zeros of $\bm{m}$ to correspond to the largest magnitude elements of $\bm{\theta}$ \cite{Lin2020Dynamic,han2015learning,guo2016dynamic}. We model different choices of $s$ amongst $\braces{s_1,\cdots,s_{K_s}}$ with the RV 
\begin{align}
\func{\bm{m}}{\bm{\pi_s}} = \sum_{k=1}^{K_s} z_s[k] \bm{m_k}, \bm{z_s} \sim \func{\text{Cat}}{\bm{\pi_s}}, \norm{\bm{m_k}}_0 / \vert \bm{m_k} \vert = s_k.
\end{align}

\textbf{Quantization} Uniformly quantized tensors are given by
\begin{align}\label{eq:bitwidth search}
    \func{Q}{\bm{\theta},b,r} = \begin{cases} 
      d \times \func{\text{round}}{\frac{\func{\text{clip}}{\bm{\theta},-r,r}}{d}}, \; \; r > 0, \; \; d = r / (2^{b-1}-1) & b > 1 \\
      \func{\text{sign}}{\bm{\theta}} & b = 1 
   \end{cases}
\end{align}
where $r$ is the quantization range and $b$ the bitwidth. We model different choices of $b \in \braces{b_1,\cdots,b_{K_q}}$ as the RV $\func{q}{\bm{\pi_q}} = \sum_{k=1}^{K_q} z_q[k] \func{Q}{\bm{\theta},b_k,r_k}, \bm{z_q} \sim \func{\text{Cat}}{\bm{\pi_q}},$
where each bitwidth $b_k$ is parameterized by its own range $r_k$. 

\textbf{Sparsity and quantization} Sparse, quantized tensors are modeled as $
    Q(\bm{\theta},b,r) \odot \bm{m}.
$
We find setting the non-zero values of $\bm{m}$ based on $\bm{\theta}$ superior to using $Q(\bm{\theta},b,r)$, which discards information about the relative magnitude of weights during the quantization process. To model tensors whose sparsity level and bitwidth must be chosen, we define the RV $q(\bm{\pi_q}) \odot \bm{m}(\bm{\pi_s})$. 
\begin{wrapfigure}{r}{0.5\textwidth}
\centering
% \vspace{-2em}
\includegraphics[width=0.95\linewidth]{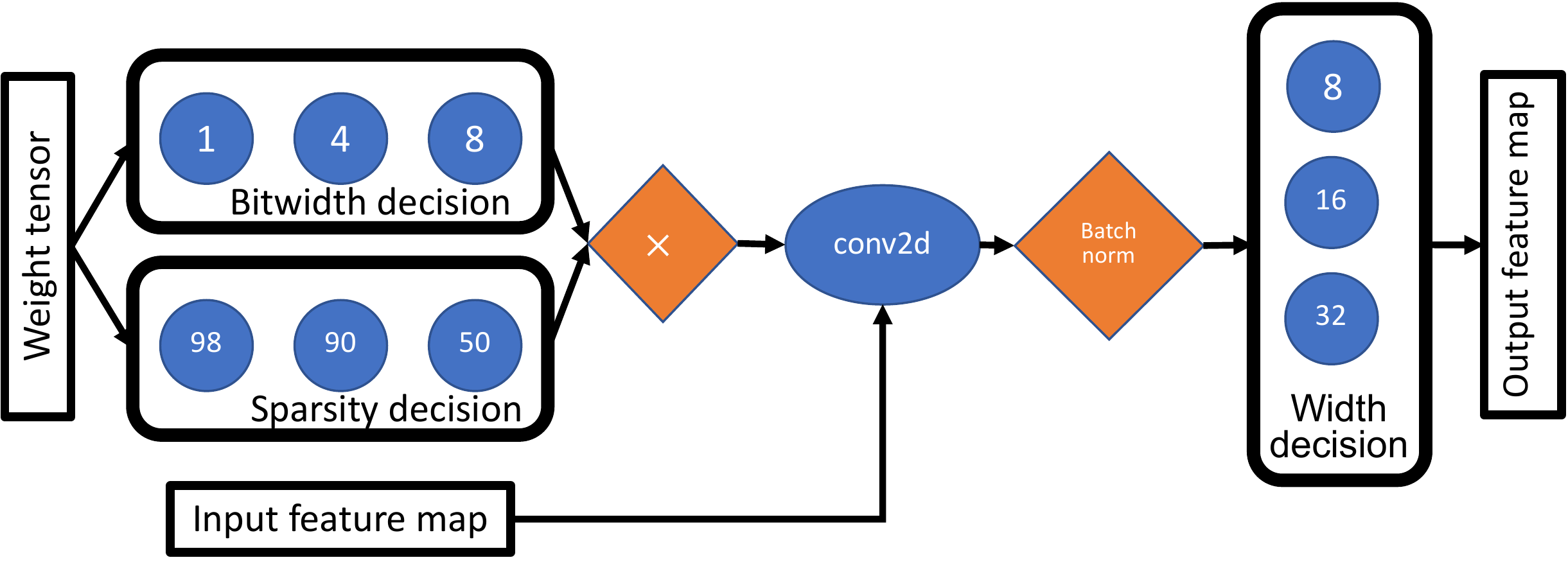}
\caption{\texttt{UDC} search space overview.}
% \vspace{-1em}
\label{fig:overview}
\end{wrapfigure}
\textbf{Operator selection}
We frame the choice over different layer operators as the RV $\sum_{k=1}^{K_f} z_f[k] f_k(\bm{x},\bm{\theta}_k), \bm{z_f} \sim \func{\text{Cat}}{\bm{\pi_f}},$
where each operator has a (possibly) different functional form $f_k(\cdot)$ parameterized by its own weights $\bm{\theta}_k$. By including identity as a candidate operation, we can also model varying NN depth.

%\textbf{Search Space Summary} 
Fig. \ref{fig:overview} shows the layer-level search space, excluding operator selection for brevity. 
Weight sharing is maximized by using the same $\bm{\theta}$ in the sparsity, bitwidth, and width decisions. The output of the layer in Fig. \ref{fig:overview} is modelled as the RV $\func{f}{\bm{x},\bm{q(\pi_q)} \odot \bm{m(\pi_s)}} \odot \bm{w(\pi_w)}$.

%\vspace{-0.5em}
\subsection{Computing layer storage size}
%\vspace{-0.5em}
%In order to solve 
Satisfying \eqref{P1}-\eqref{P2} necessitates a quantitative measure of storage size,
%which accounts 
accounting for data compression. 
For a given layer, the storage size achievable by prefix-free compression is lower-bounded by the weight (empirical) entropy, $H(\cdot)$, times the number of weight elements \cite{cover1999elements}. The entropy bound can, in turn, be bounded by (Appendix \ref{section:derivation of entropy bound}):
\begin{align}
\func{H}{Q(\bm{\theta},b,r) \odot \bm{m} } \times \norm{\bm{w}}_0   \leq (b + \left( -s \log_2 s - (1-s)\log_2(1-s) \right)) \times \norm{\bm{w}}_0  \label{eq:practical entropy}.
\end{align}
The advantage of the right hand side (RHS) of \eqref{eq:practical entropy} over the left hand side (LHS) is that it can be computed cheaply without processing $\bm{\theta}$. Using the LHS requires computing the empirical entropy, which is expensive for large $\bm{\theta}$. 
Moreover, combining the LHS with a gradient-based optimizer requires gradient approximation, since the empirical entropy is not differentiable \cite{theis2017lossy}. We refer to the RHS of \eqref{eq:practical entropy} as the compressed tensor size and use it as our measure of layer size. Sec.
\ref{section:ablations} confirms that it is achievable with both an arithmetic encoder and Vela. Let ${\epsilon}(s,b,\rho)$ be a given layer's storage size (the RHS of \eqref{eq:practical entropy}) as a function of sparsity $s$, bitwidth $b$, and non-zero channel fraction $\rho$. When $(s,b,\rho)$ must be chosen, let $\func{\mathcal{\epsilon}}{\sum_{k}^{K_s} z_s[k] s_k, \sum_k^{K_q} z_q[k] b_k, \sum_k^{K_w} z_w[k] \rho_k}$ be the storage size. For the entire NN, we sum the storage size of all layers and denote the result $\func{\mathcal{E}}{\braces{\bm{z}}}$, where $\braces{\bm{z}}$ is short-hand for the set of all decision variables. 
%\vspace{-2mm}
\vspace{-0.5em}
\section{Proposed DNAS algorithm}
\label{section:dnas}
%\vspace{-1em}
Our optimization objective is:%\vspace{-1em}
% \small
\begin{equation}\label{eq:objective}
\begin{aligned}
    \argmin_{\braces{\bm{\pi},\bm{\theta},\bm{m},r}} &E_{\braces{\bm{z}},\mathcal{D}} \left[ \func{L}{\braces{\bm{z},\bm{\theta}},\mathcal{D}} \right] 
    \textrm{ s.t. } \func{\mathcal{E}}{\braces{\gamma(\bm{\pi})}} = {e}^*, \; \gamma(\bm{\pi}) = \func{\text{onehot}}{\argmax_k \pi[k]}
\end{aligned}
\end{equation}
% \normalsize
where $L(\cdot)$ is a task loss, e.g. cross-entropy, $\mathcal{D}$ is the training data, and $e^*$ is the target model size. The constraint in \eqref{eq:objective} stems from the two-stage process typical in DNAS: 1) Optimize $\braces{\bm{\pi},\bm{\theta}}$ and extract the most likely configuration $\braces{\gamma(\bm{\pi})}$, 2) Train and deploy the result \cite{liu2018darts,dong2019network,dong2019searching,jin2019rc,wan2020fbnetv2,wu2019fbnet}.
While prior works use the constraint \cite{dong2019network,dong2019searching,jin2019rc,wan2020fbnetv2,gong2019mixed}
\begin{align}\label{eq:constraint in expectation}
    E_{\braces{\bm{z}}}\left[\func{\mathcal{E}}{\braces{\bm{z}}}\right] = e^*,
\end{align}
we constrain the \textit{most likely} configuration since this is what is actually deployed. Even if \eqref{eq:constraint in expectation} is satisfied, it is not guaranteed that $\braces{\gamma(\bm{\pi})}$, the deployed model, satisfies the constraint.

\textbf{Gradient-based optimization} 
Solving \eqref{eq:objective} using stochastic gradient descent (SGD) requires: 1) ensuring that the constraint is met, 2) differentiating with respect to $\braces{\bm{\theta},\bm{\pi}}$. To deal with the constraint and avoid dealing with the non-differentiable $\gamma(\bm{\pi})$, we modify \eqref{eq:objective} to
\begin{align}\label{eq:regularized objective}
    \argmin_{\braces{\bm{\pi},\bm{\theta},\bm{m},r}} & \underbrace{E_{\braces{\bm{z}},\mathcal{D}} \left[ \func{L}{\braces{\bm{z},\bm{\theta}},\mathcal{D}} \right]}_{\mathcal{L}_{\text{task}}} + \lambda \underbrace{ E_{\braces{\bm{z}}} \left[ \vert \func{\mathcal{E}}{\braces{\bm{z}}} - e^* \vert  \right]}_{\mathcal{L}_{\mathcal{E}}^z}.
\end{align}
While \regularizer{z} has a different form than the constraint in \eqref{eq:objective}, it actually represents a much stronger constraint and its minimization implies the constraint is met.
\begin{lemma}\label{lemma:regularizer-constraint}
If \regularizer{z}$=0$, then for any sample of $\braces{\bm{z}}$, denoted $\braces{\bm{z}^s}$, with non-zero probability, $\func{\mathcal{E}}{\braces{\bm{z}^s}} = e^*$ and $\func{\mathcal{E}}{\braces{\gamma(\bm{\pi})}} = e^*$.
\end{lemma}
\regularizer{z} differs from \eqref{eq:constraint in expectation} because it penalizes any configuration which violates the constraint, whereas \eqref{eq:constraint in expectation} penalizes $\braces{\bm{\pi}}$ only if the expected storage size violates the constraint. 

The derivative of \eqref{eq:regularized objective} with respect to (w.r.t.) $\braces{\bm{\theta}}$ can be approximated using a Monte-Carlo (MC) approximation of the expectation and applying standard automatic differentiation. The derivative w.r.t. $\braces{\bm{\pi}}$ is more complex since the expectations in \eqref{eq:regularized objective} depend on $\braces{\bm{\pi}}$. A popular solution is to use a biased but differentiable approximation of $\bm{z}$, given by the Gumbel-softmax distribution: $\hat{\bm{z}} = \func{\textrm{softmax}}{\frac{\log \bm{\pi} + \bm{g}}{\tau}}, g[k] \sim \func{\textrm{Gumbel}}{0,1}, \tau > 0$ \cite{jang2016categorical,dong2019network,dong2019searching}. As $\tau \rightarrow 0$, $\hat{\bm{z}}$ approaches $\bm{z}$ in distribution, while the variance of any gradient estimator which uses $\hat{\bm{z}}$ increases (\cite{paulus2020rao} and Fig. \ref{fig:max_var}). As a result, the common practice is to anneal $\tau$ from a high to a low value throughout the search. Using $\hat{\bm{z}}$ can lead to co-adaptation of search space options, which is undesirable but solvable in practice (Appendix \ref{section:avoiding co-adaptation}).

\begin{wraptable}{r}{0.5\textwidth}
% \vspace{-1em}
\caption{Values of $\mathcal{L}_{\mathcal{E}}^{\hat{z}}$ for different values of $\tau, \xi \text{(see \eqref{eq:projection_set})}, \vartheta \text{(Sec. \ref{subsection:rejection sampling})}$. The search space is based on MBNetV2 (Sec. \ref{section:results}) and $\func{\mathcal{E}}{\braces{\gamma(\bm{\pi})}} = e^*$ in all cases.
}
\label{table:tau}
\centering
\resizebox{0.5\textwidth}{!}{%
\begin{tabular}{ccccc}
\toprule
      & \textbf{Vanilla}              & \textbf{Projection}                 & \multicolumn{2}{c}{\textbf{Projection \& Rejection Sampling} } \\
      &  \textbf{DNAS}           & $\xi=0.5, \vartheta = 0$ & $\xi=0.5, \vartheta = 0.5$ &  $\xi=0.5, \vartheta = 0.99$\\
\midrule
$\tau=0.66$ & 0.04 & 0.33 & 0.27 & 0.18 \\
$\tau=10$ & 0.53 & 0.61  & 0.6 & 0.59 \\
\bottomrule
\end{tabular}}
%\vspace{-1em}
\end{wraptable}
\textbf{Gumbel-softmax and over-regularization} 
We observe two issues with annealing $\tau$: 1) increased gradient variance at low $\tau$, coupled with a complex search space, causes issues for SGD, 2) when we replace $\bm{z}$ with $\hat{\bm{z}}$ in \regularizer{z}, with the result \regularizerhat{z}, the regularizer becomes artificially inflated. 
To understand the impact of $\tau$ on \regularizerhat{{z}}, we evaluate \regularizerhat{{z}} for different values of $\tau$, setting $\bm{\pi}$ such that $\func{\mathcal{E}}{\braces{\gamma(\bm{\pi})}} = {e}^*$. The results are presented in Table \ref{table:tau}, col. 1 and show that increasing $\tau$ increases \regularizerhat{{z}}, i.e. the relative impact of \regularizerhat{{z}} on \eqref{eq:regularized objective} depends on $\tau$. As such, we seek to keep $\tau$ low, while minimizing gradient variance. Our solution is to use multiple samples of $\braces{\hat{\bm{z}}}$ in the MC approximation of \eqref{eq:regularized objective}. To maintain the same computational cost as the single MC sample case, we divide the number of optimization steps by the number of samples. As well as reducing gradient variance, our strategy has two additional practical benefits: 1) trivial extension to multi-GPU systems, since each GPU can run its own MC sample and gradient computation, 
2) the overheads of computing the gradient are amortized across the MC samples. We observe a $5.3\times$ speed-up when going from $1$ MC sample to $32$ (Table \ref{table:ablation}, col. 6, row 4 vs. row 8.) 
%\vspace{-1em}
\subsection{Exploration-exploitation}
\label{subsection:exploration-exploitation}
%\vspace{-1em}
When solving \eqref{eq:regularized objective}, the goal is to 
%balance exploration and exploitation, giving 
explore as many configurations as possible (exploration), while still training each configuration for a meaningful number of steps (exploitation). 
Ideally, the search algorithm should gradually move from exploration to exploitation. We propose to explicitly control the exploration-exploitation trade-off by projecting $\bm{\pi}$ onto the set 
\begin{align}\label{eq:projection_set}
\mathcal{S} = \braces{\bm{\pi}: \max_k {\pi}[k] \leq 1/\vert \bm{\pi} \vert + \xi^t}
\end{align}
after each SGD step, where $\xi^t$ is the upperbound on $\bm{\pi}$ at step $t$. Setting $\xi^t = 0$ constrains $\bm{\pi}$ to parameterize a uniform distribution and represents maximal exploration. Setting $\xi^t = 1 - 1/\vert \bm{\pi} \vert$ removes the constraint on $\bm{\pi}$, allowing the optimizer to enter full exploitation. We define the projection operator as $P_{\mathcal{S}}(\bm{\pi}) = \func{\text{softmax}}{\log \bm{\pi} / T^*}$, $T^* = \argmin_{T} \sum_k^{\vert \bm{\pi} \vert} \func{\text{max}}{P_{\mathcal{S}}(\bm{\pi})[k] - (1 / \vert \bm{\pi} \vert + \xi^t),0}$, 
which we solve numerically. We choose this form for $P_{\mathcal{S}}(\bm{\pi})$ because of its simplicity and because the relative ordering of options between $\bm{\pi}$ and $P_{\mathcal{S}}(\bm{\pi})$ does not change. While we find it necessary to enforce exploration by projecting $\bm{\pi}$ onto $\mathcal{S}$, \regularizer{z} implicitly promotes exploitation. 
\begin{lemma}\label{lemma:one hot}
Let \regularizer{z} $=0$ and $\bm{z}_j$ be the $j$'th decision variable. Let there be no decision for which two of its options have the same cost, i.e. for two configurations $\braces{\bm{z}^s}$ and $\braces{\bm{z}^{s'}}$ such that $\bm{z}_j^s \neq \bm{z}_j^{s'}$ and $\bm{z}_k^s = \bm{z}_k^{s'} \forall k \neq j$, we have \regularizer{{z^s}} $\neq$ \regularizer{{z^{s'}}}. Then each $\bm{\pi}_j$ must be one-hot.
\end{lemma}
The assumption in Lemma \ref{lemma:one hot} that no decision has two options with the same storage cost is satisfied for a typical compressible model search space.
%\vspace{-1em}
\subsection{Combating over-regularization with rejection sampling}
\label{subsection:rejection sampling}
%\vspace{-1em}
Projecting $\bm{\pi}$ onto $\mathcal{S}$ enables explicit control over the exploration-exploitation dynamics, but 
%it 
also inflates \regularizerhat{z}. Table \ref{table:tau}, col. 2 shows that setting $\xi^t = 0.5$ increase \regularizerhat{z} dramatically. By forcing $\bm{\pi}$ to be closer to uniform, the number of configurations $\braces{\hat{\bm{z}}}$ with non-zero probability increases, such that the probability of a randomly drawn configuration violating the constraint also increases. Inflating \regularizerhat{z} forces the optimizer to focus less on \task{}, leading to solutions which meet the constraint but perform poorly on the target task. \regularizerhat{z} increases when the properties of $\braces{\bm{\pi}}$ change because it depends on all possible configurations $\braces{\hat{\bm{z}}}$ instead of the most likely one, i.e. $\braces{\gamma(\bm{\pi})}$. To motivate the remedy, observe that while not all samples of $\hat{\bm{z}}$, denoted $\hat{\bm{z}}^s$, satisfy $\gamma(\hat{\bm{z}}^s) = \gamma(\bm{\bm{\pi}})$, some do. Indeed, \regularizerhat{z} would be $0$ if it was evaluated over those samples that satisfy $\gamma(\hat{\bm{z}}^s) = \gamma(\bm{\bm{\pi}})$, assuming $\func{\mathcal{E}}{\braces{\gamma(\bm{\bm{\pi}})}} = e^*$ (and $\kappa=1$ in Appendix \ref{section:avoiding co-adaptation}  \eqref{eq:ste}). We refer to samples generated in this manner as $\tilde{\bm{z}}^s$ and they correspond to a RV whose distribution is different from $\hat{\bm{z}}$, but still depends on $\bm{\pi}$ and can therefore be used to generate gradients to $\bm{\pi}$ from \regularizer{{\tilde{z}}}. Alg. \ref{alg:rejection sampling} shows how to generate $\tilde{\bm{z}}^s$. Replacing $\hat{\bm{z}}$ with $\tilde{\bm{z}}$ for all decisions negates the effects of controlling $\bm{\pi}$ through projection onto $\mathcal{S}$. Therefore, we use $\tilde{\bm{z}}$ for a given decision with probability $\vartheta$ and $\hat{\bm{z}}$ otherwise.
\begin{minipage}{0.46\textwidth}
% \vspace{-1em}
\begin{algorithm}[H]
\caption{Rejection Sampling}
\label{alg:rejection sampling}

\begin{algorithmic}[1]
\State Sample $\hat{\bm{z}}^s \sim \func{p}{\hat{\bm{z}}}$, $1 \leq s \leq S$
\State $\tilde{s} = 1$, $k^* = \argmax_k \bm{\pi}[k]$

\For{$1 \leq s \leq S$}
\State $\hat{k} = \argmax_k \hat{z}^s[k]$
\If{$\hat{k} = k^*$}
\State $\tilde{\bm{z}}^{\tilde{\bm{s}}} = \hat{\bm{z}}^s$, $\tilde{s} = \tilde{s} + 1$
\EndIf
\EndFor
\State $\tilde{\bm{z}} = \frac{1}{\tilde{s}-1} \sum_{s'=1}^{\tilde{s}-1} \tilde{\bm{z}}^{s'}$ \Comment{Average}
\State \textbf{return} $\tilde{z}$
\end{algorithmic}
\end{algorithm}
\end{minipage}
\hfill
\begin{minipage}{0.46\textwidth}
% \vspace{-1em}
\begin{algorithm}[H]
\caption{Complete \texttt{UDC} Algorithm}
\label{algorithm:complete}
\begin{algorithmic}[1]
\For{$1 \leq t \leq t_{\text{max}}$}
\For{$1 \leq s \leq S$}
\State Generate MC sample $s$ for decision $j$ using $\hat{\bm{z}}_j$ w.p. $\vartheta$ and $\tilde{\bm{z}}_j$ else
\EndFor
\State Update $\braces{\bm{m}}$ if $t \text{ modulo } 16 = 0$
\State Take SGD step on \eqref{eq:regularized objective}
\State $\bm{\pi} \leftarrow P_{\mathcal{S}}(\bm{\pi})$ \Comment{Projection}
\EndFor
\State \textbf{return} $\braces{\bm{\pi}}$
\end{algorithmic}
\end{algorithm}
\end{minipage}

Using $\tilde{\bm{z}}$ has two major benefits. First, even when exploration is enforced by projecting $\bm{\pi}$ onto $\mathcal{S}$, \regularizer{{\tilde{z}}} is not inflated. 
Table \ref{table:tau}, col.s 3-4 show how $\vartheta > 0$ brings down \regularizerhat{z} for the same underlying $\braces{\bm{\pi}}$. 
Second, mixing $\hat{\bm{z}}$ and $\tilde{\bm{z}}$ gives the flexibility of being in exploration for some decisions and exploitation for others. By randomly choosing which decisions use $\hat{\bm{z}}$ and which use $\tilde{\bm{z}}$, we prevent greedy behavior whereby a given decision enters exploitation and never returns to exploration. The complete \texttt{UDC} algorithm is summarized in Alg. \ref{algorithm:complete}. 
% \vspace{-1em}
\section{Training sparse, quantized models}
\label{sec:training sparse quantized models}
%\vspace{-1em}
DNAS is typically a two-stage process: a search to find the model architecture is followed by finetuning to find optimal weights. 
However, we find that the second stage yields poor results when training sparse, quantized models from scratch~\cite{zhou2017incremental}. 
\texttt{UDC} employs a three-stage finetuning process: 
\textbf{Stage 1:} initialize $\braces{\bm{\theta}}$ and train with quantization enabled but unstructured pruning disabled, 
\textbf{Stage 2:} enable unstructured pruning gradually, 
\textbf{Stage 3:} train with both quantization and unstructured pruning enabled. 
To counter the training challenges induced by quantizing and pruning $\bm{\theta}$, we employ several known techniques, with one slight modification. For quantization, we use a variant of \cite{stock2021training} where each weight is quantized with probability $\alpha$ during the forward pass, such that the weights used for training are $Q(\bm{\theta},b,r) \odot \bm{h} + \func{\text{clip}}{\bm{\theta},-r,r} \odot (\bm{1}-\bm{h}), h[i,j,c_{in},c_{out}] \sim \func{\text{Bernoulli}}{\alpha}$. Unlike \texttt{UDC}, \cite{stock2021training} uses $\bm{\theta}$ in the second term. We observe that the range of $\bm{\theta}$ can differ significantly from that of $Q(\bm{\theta},b,r)$, especially when unstructured pruning is applied during the learning process, so we clip $\bm{\theta}$ to the same range as $Q(\bm{\theta},b,r)$. For unstructured pruning, we gradually anneal the pruning rate from $0\%$ to $100\%$ of the target rate during stage 2 \cite{zhu2017prune,Lin2020Dynamic}.

\begin{figure*}
    \centering
    \begin{subfigure}[t]{0.33\linewidth}
        \centering
        \includegraphics[scale=0.21]{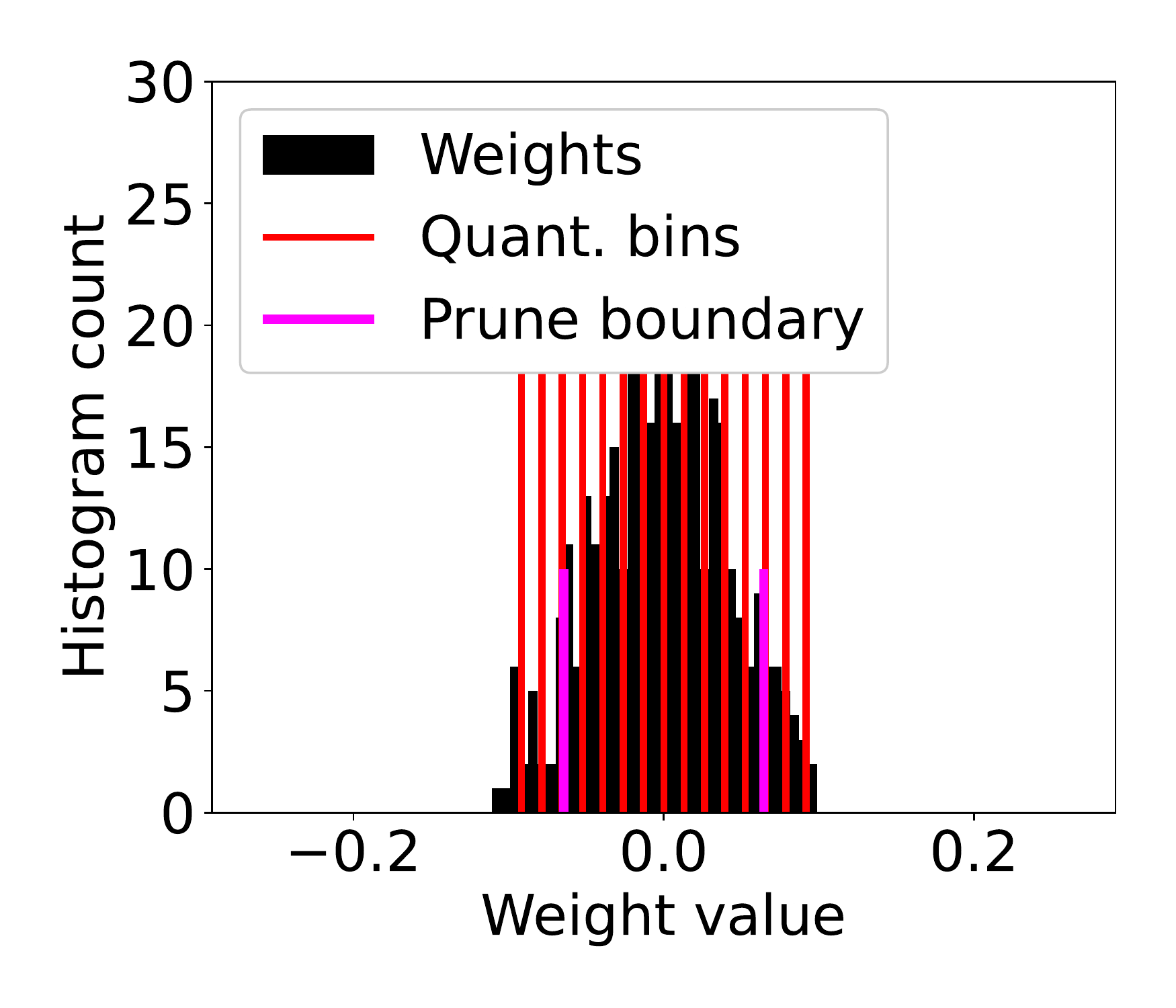}
        % \vspace{-1em}
        \caption{Baseline, no retraining}
        \label{fig:baseline representation}
    \end{subfigure}%
    ~ 
    \begin{subfigure}[t]{0.33\linewidth}
        \centering
        \includegraphics[scale=0.21]{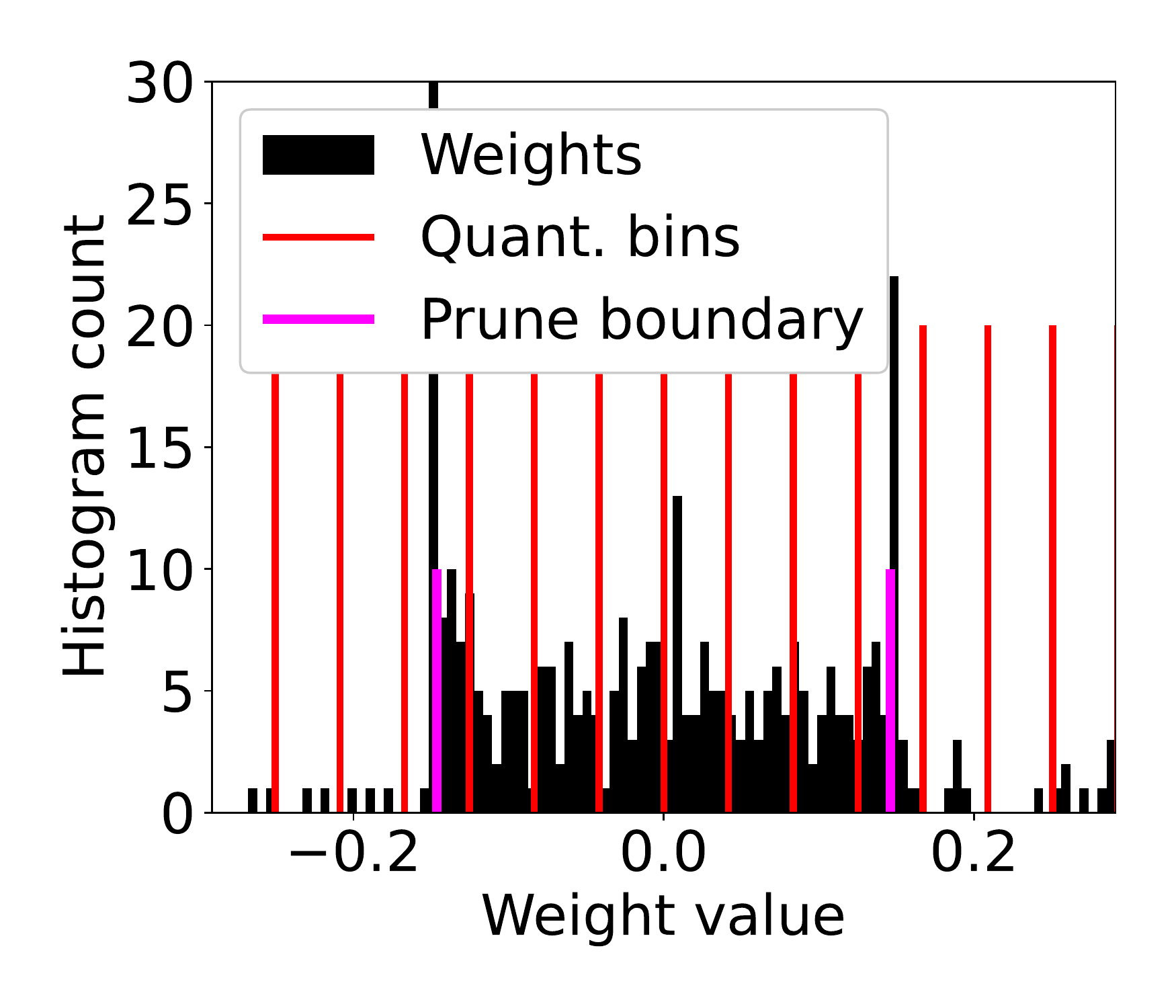}
        % \vspace{-1em}
        \caption{Baseline, retrained}
        \label{fig:baseline representation retrained}
    \end{subfigure}%
    ~ 
    \begin{subfigure}[t]{0.33\linewidth}
        \centering
        \includegraphics[scale=0.21]{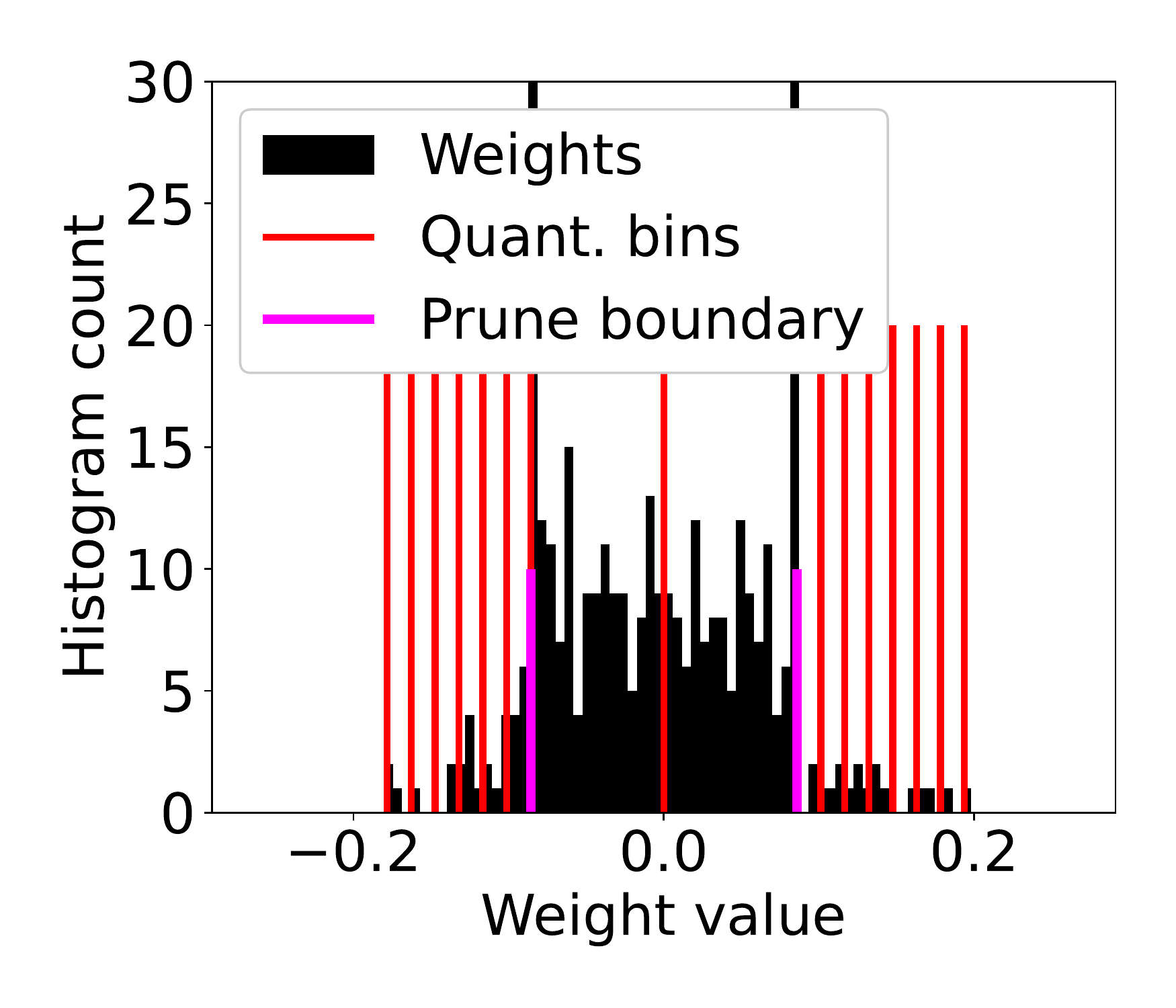}
        % \vspace{-1em}
        \caption{Proposed, retrained}
        \label{fig:proposed representation}
    \end{subfigure}
\caption{Histogram of weights from a single pruned and quantized layer in the ImageNet experiment under different conditions. Weight norm $\left \Vert \theta \right \Vert_2^2$: (\ref{fig:baseline representation}) 1.7e3, (\ref{fig:baseline representation retrained}) 7.9e3, (\ref{fig:proposed representation}) 3.7e3.}
%\vspace{-1em}
\label{fig:number representation}
\end{figure*}

\begin{figure}
        \centering
    \begin{subfigure}[t]{0.24\linewidth}
        \centering
        \includegraphics[width=0.95\linewidth]{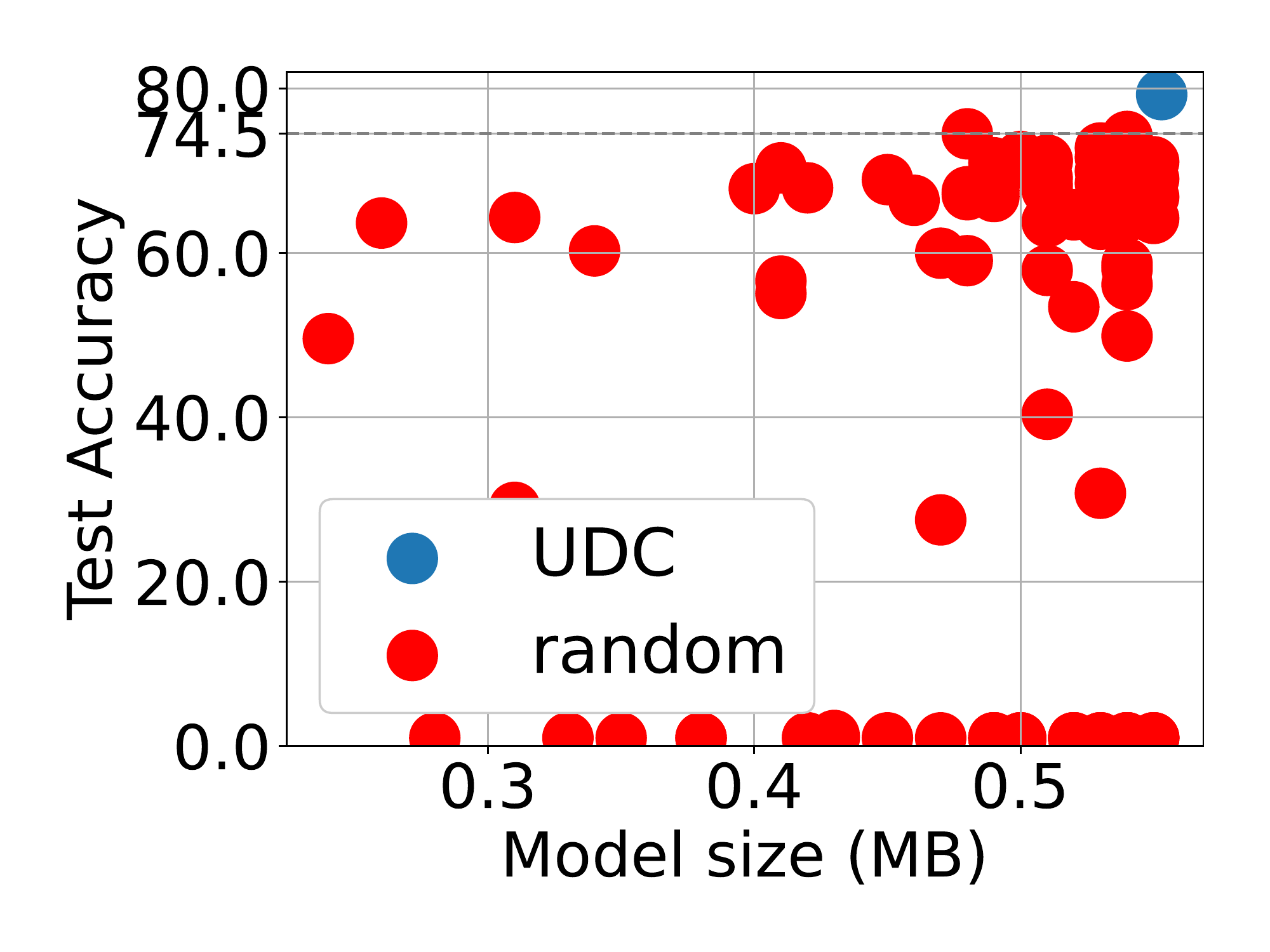}
        % \vspace{-0.5em}
        \caption{}
        \label{fig:random baseline scatter}
    \end{subfigure}%
    ~
    \begin{subfigure}[t]{0.24\linewidth}
        \centering
        \includegraphics[width=0.95\linewidth]{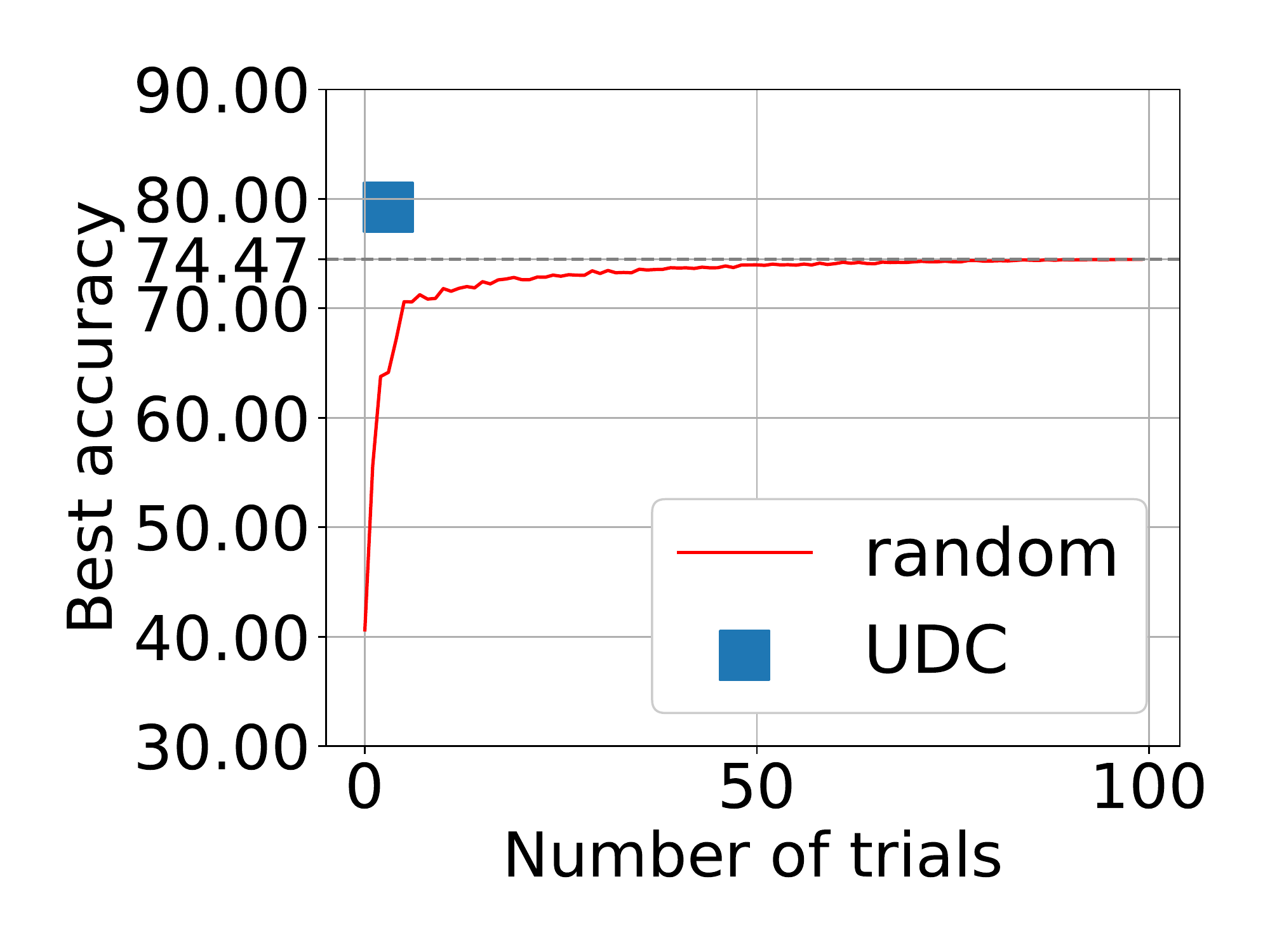}
        % \vspace{-0.5em}
        \caption{}
        \label{fig:random baseline bootstrap}
    \end{subfigure}%
    ~
    \begin{subfigure}[t]{0.24\linewidth}
        \centering
    \includegraphics[width=0.95\linewidth]{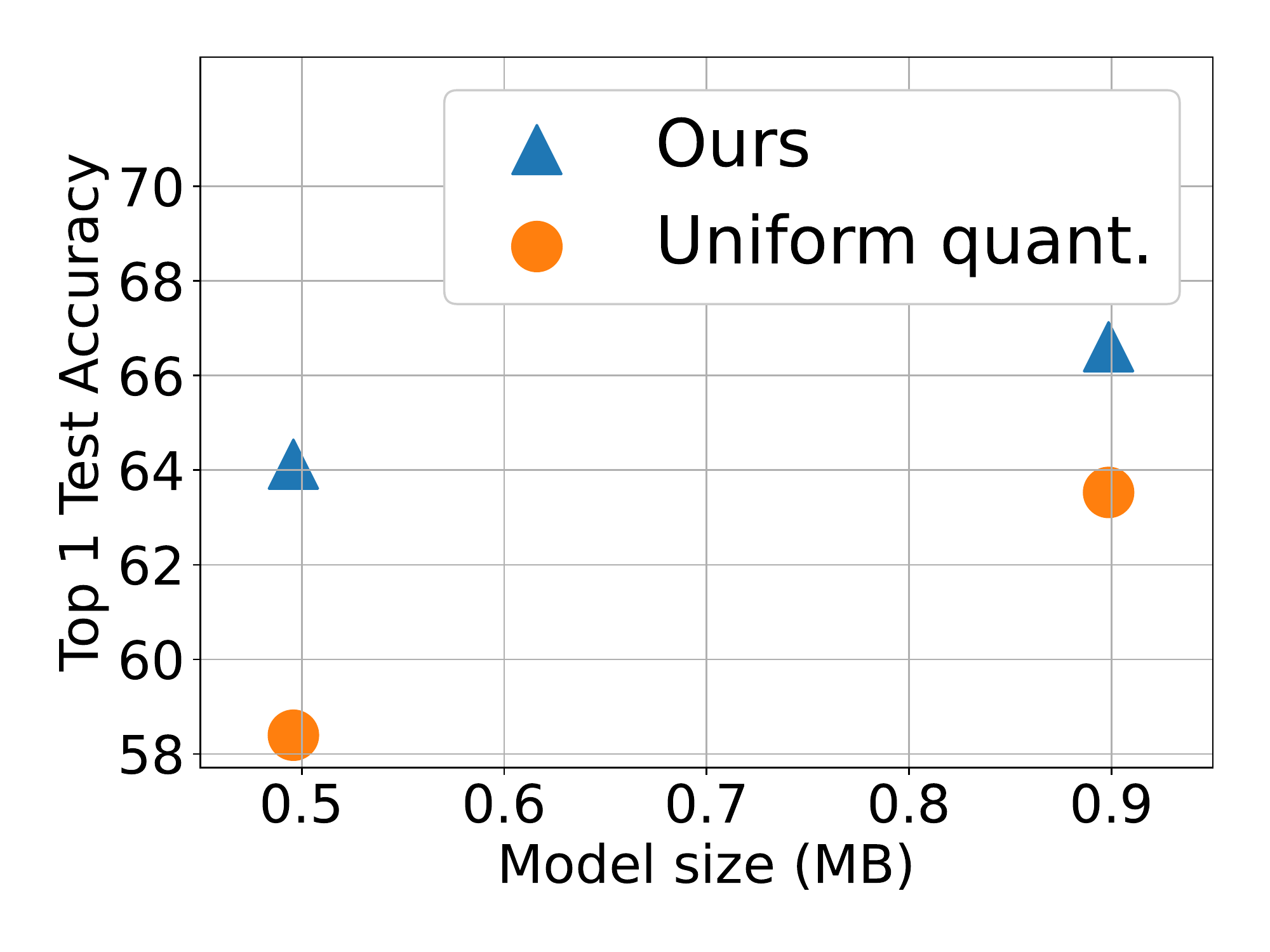}
    % \vspace{-0.5em}
    \caption{}
    \label{fig:number format}
    \end{subfigure}%
    ~
    \begin{subfigure}[t]{0.24\linewidth}
        \centering
    \includegraphics[width=0.95\linewidth]{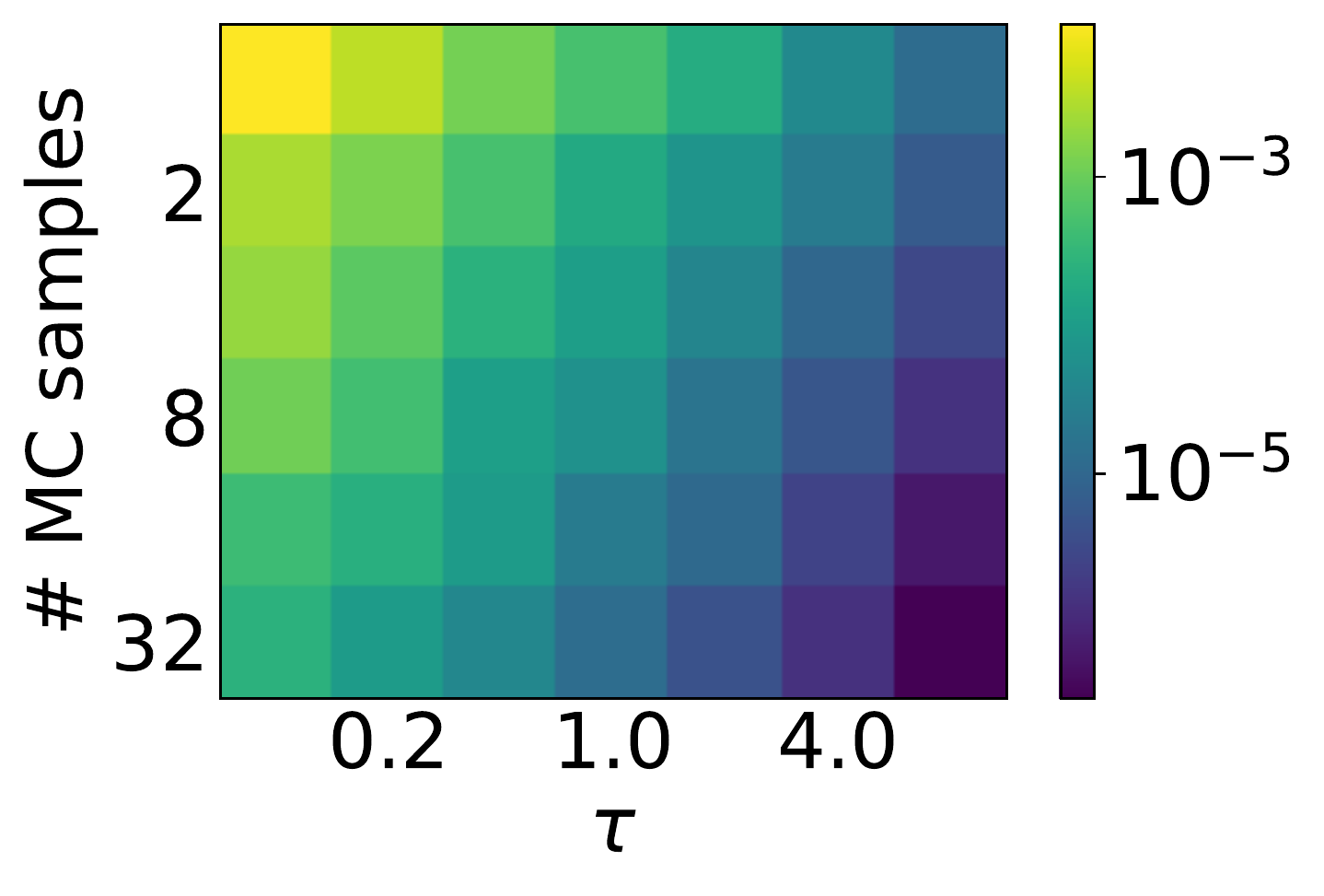}
    % \vspace{-0.5em}
    \caption{}
    \label{fig:max_var}
    \end{subfigure}
\caption{\ref{fig:random baseline scatter}-\ref{fig:random baseline bootstrap}: Models found by random search. \ref{fig:number format}: Benefit of novel number format on ImageNet. \ref{fig:max_var}: Maximum gradient variance over elements of $\braces{\bm{\pi}}$ for CIFAR100 experiment.}
%\vspace{-2.3em}
\label{figu:random baseline}
\end{figure}

\textbf{Weight numerical representation}
To understand the challenge of training sparse, quantized NNs, consider a single layer of pretrained weights, with histogram in Fig. \ref{fig:baseline representation}.
The quantization bins ($b=4$) are shown in red and the pruning boundary in purple, with everything between the purple lines mapped to $0$. The sparse quantization problem is clearly apparent here: $8/14$ of the non-zero quantization bins are unused because they fall \textit{inside} the pruning boundary. 
If we now train with weight sparsity and quantization constraints, the weight distribution, quantization bins, and pruning zone adjust (Fig. \ref{fig:baseline representation retrained}). The optimizer flattens the weight distribution to use more quantization bins, resulting in increased weight range. To quantify the weight growth, we report the norm of the NN weights before and after training in Fig. \ref{fig:number representation}, showing an increase of over $4.6\times$. 
The rate at which NNs can be trained, known as the effective learning rate, is inversely proportional to the weight norm \cite{arora2018theoretical}. Therefore, the interaction of sparsity and quantization cause weight norm inflation, which decreases the effective learning rate, reducing NN performance. We propose a different weight representation and only quantize the range beyond the pruning boundary (Fig. \ref{fig:proposed representation}), using the quantization operator $\hat{Q}(\bm{\theta},b,r,\beta) = Q(\bm{\theta} - \text{sign}(\bm{\theta}) \beta,b,r) + \text{sign}(\bm{\theta}) \beta $ for $b > 1, \beta \in \mathbb{R}$. We set $\beta$ to be the largest pruned value of $\bm{\theta}$. Training the sparse, quantized NN with the proposed number representation leads to much smaller weight norm (only $2.2\times$ growth over the pretrained weights), which makes training easier and accuracy higher (Fig. \ref{fig:number format}).

\textbf{Deployment with integer math}
NPU/MCU HW platforms typically only support integer operations, which are cheaper than floating point. There are at least two ways of deploying NNs quantized using the proposed approach on such HW. 
Firstly, convolution can be decomposed into $f(\bm{x},Q(\bm{\theta}-\text{sign}(\bm{\theta}) \beta,b,r)) + \beta f(\bm{x},\text{sign}(\bm{\theta}))$. 
Both terms can be calculated using only integers, but since $\text{sign}(\bm{\theta})$ is a 1-bit tensor, the second term does not require any multiplications. 
Secondly, the NN can be trained using the proposed number format, then deployed with $\func{Q}{\hat{Q}(\bm{\theta},b,r,\beta),b^*,r'}$, such that the deployed NN is uniformly quantized, with $b^*=8$ to match the datatype supported in MCUs/NPUs~\cite{tflitemicro,utensor,ethos-u55}.
The advantage of the latter approach is that the training benefits from the expressivity of the proposed number format, while its deployment uses a standard data type.
We find that this approach with $b^*=8$ does not incur an accuracy loss on ImageNet (Fig. \ref{fig:imagenet}).

%\vspace{-1.5em}
\section{Results}
\label{section:results}
%\vspace{-1em}

We compare \texttt{UDC} with SOTA methods on a model size vs. accuracy basis. 
All reported model sizes, including related works, use compressed size (RHS of \eqref{eq:practical entropy}), except Choi et al. \cite{choi2020learning}, who use bzip2 to compress weights and we use their reported sizes.

\begin{figure}
    \centering
    \begin{subfigure}[t]{0.33\linewidth}
         \centering
    \includegraphics[width=\linewidth]{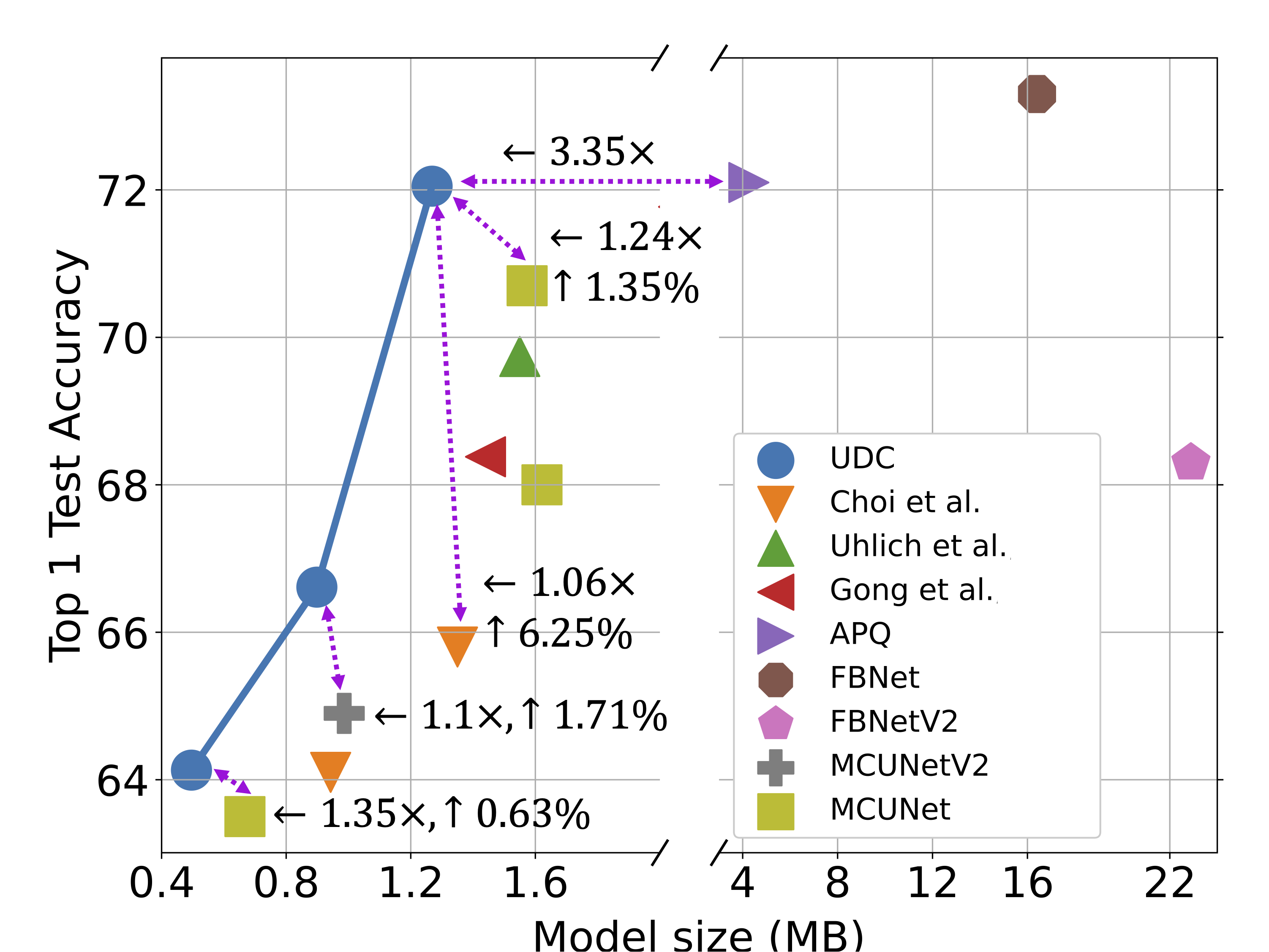}
    \caption{ImageNet classification}
    \label{fig:imagenet}
    \end{subfigure}%
    ~
    \begin{subfigure}[t]{0.33\linewidth}
         \centering
    \includegraphics[width=\linewidth]{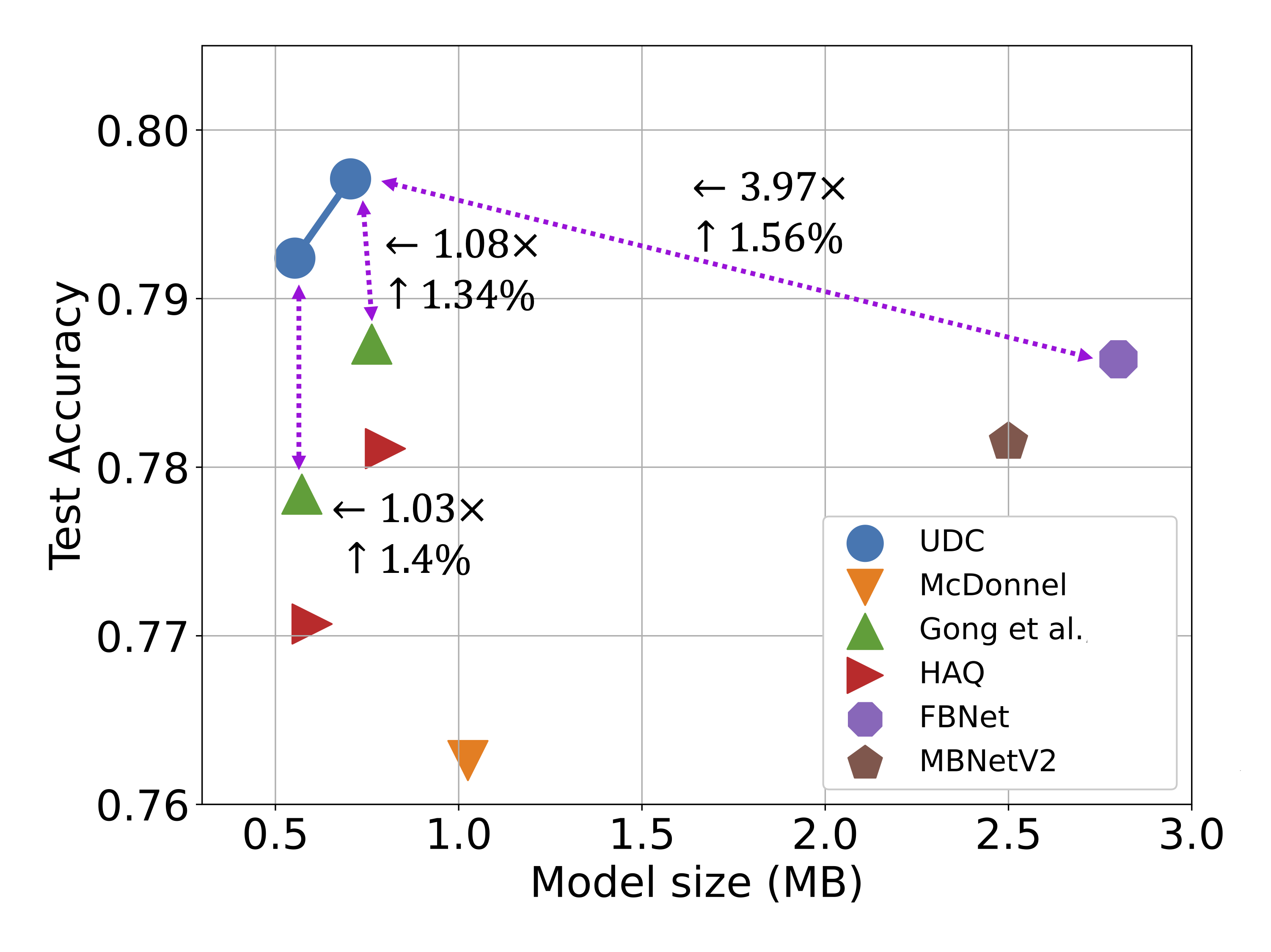}
    \caption{CIFAR100 classification}
    \label{fig:CIFAR100 results}
    \end{subfigure}%
    ~
    \begin{subfigure}[t]{0.33\linewidth}
        \centering
    \includegraphics[width=\linewidth]{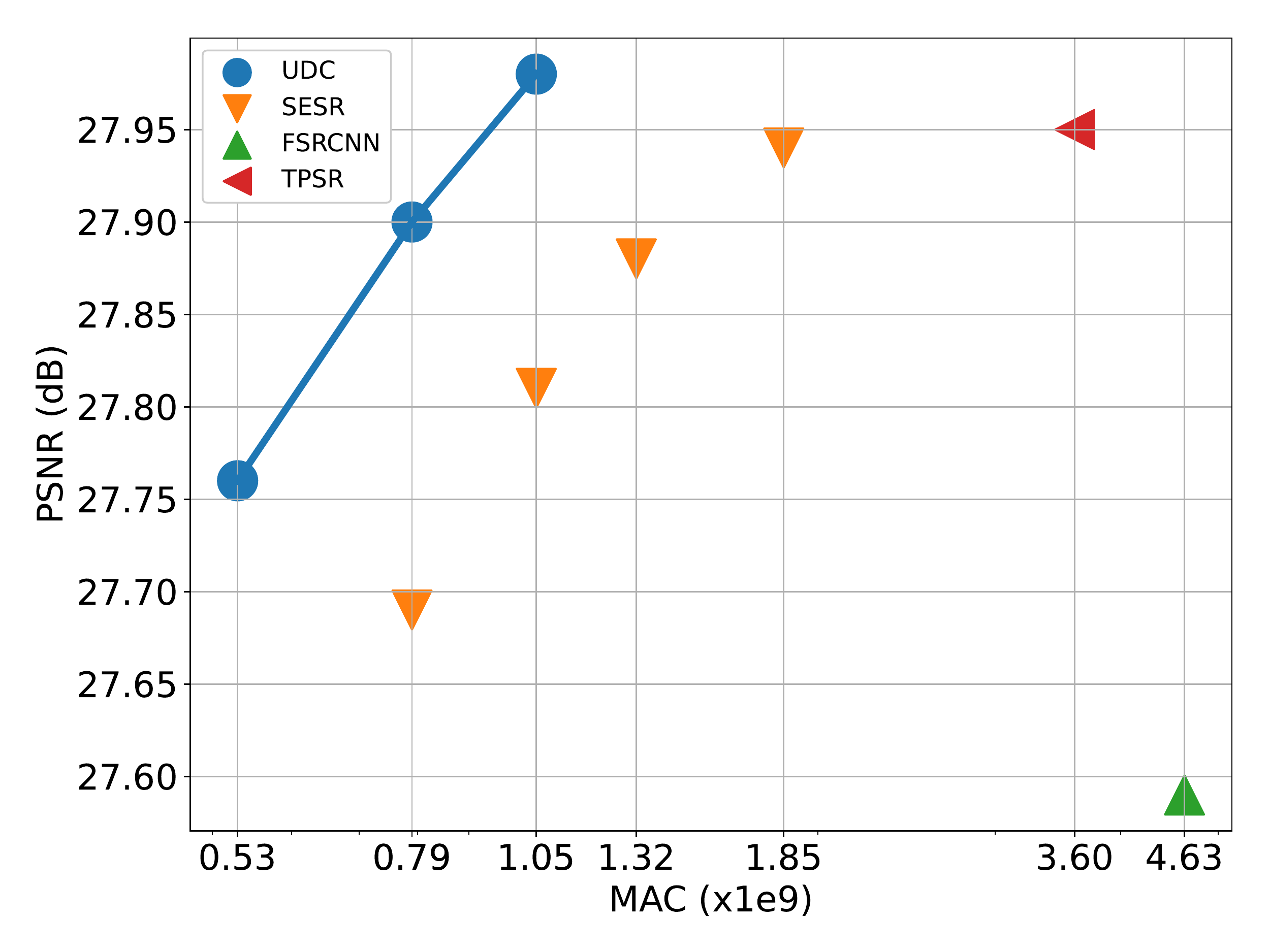}
    \caption{set14 super resolution}
    \label{fig:super resolution}
    \end{subfigure}
% \vspace{-0.5em}
\caption{\texttt{UDC} finds pareto-dominant models on image tasks.}
% \vspace{-0.8em}
\label{figu:imagenet and cifar100 and super resolution}
\end{figure}

\textbf{CIFAR100}
is an image classification task
%recognition dataset 
with $50$k training / $10$k test images, and $100$ classes. Our search space is based on the wide residual network with depth $20$ and width multiplier $1$0 \cite{zagoruyko2016wide,mcdonnell2018training}. We search over layer width (increments of $10\%$ of the orignal), bitwidth (1, 4, 8, 32), and sparsity ($1-100\%$ non-zeros, increments of $10\%$). 
We use \texttt{UDC} (
%with the 
settings 
%are 
%given
%described 
in Appendix \ref{section:cifar100 experiment settings}) 
to find and train models at two sizes: $0.55$MB and $0.7$MB. 
Fig. \ref{fig:CIFAR100 results} shows that \texttt{UDC} generates Pareto-dominant models compared with
%compares our results with
SOTA methods \cite{gong2019mixed,mcdonnell2018training,wu2019fbnet,wang2018haq,sandler2018mobilenetv2}.
%to demonstrate that UDC generates Pareto-dominant NNs.

\textbf{ImageNet} is an image 
%recognition dataset 
classification task with $1.28$M training / $50$k test images, and $1$k classes. 
Our search space is based on MBNetV2, 
%searching over 
with the same options as 
%the 
for CIFAR100,
%experiment, 
and we target $0.5$, $1$, and $1.25$ MB models (Appendix \ref{section:ImageNet experiment settings}). For the $1.25$MB experiment, we replace $3$$\times$$3$ kernels with $5$$\times$$5$ and make all layers $1.5$$\times$ wider in the baseline architecture.

In practice, the model size constraints are determined by the Flash memory size of the deployment HW platform. To be sure, $0.5-1.25$ MB Flash sizes are fairly common for commodity HW platforms \cite{stm32} and are often used in research targeting deployment on constrained HW platforms \cite{lin2020mcunet, banbury2021micronets}. As such, we targeted this range because it represents a reasonable, but extremely challenging deployment scenario.

Fig. \ref{fig:imagenet} shows that \texttt{UDC} generates pareto-dominant NNs 
%with respect to 
vs. the SOTA \cite{lin2020mcunet,choi2020learning,gong2019mixed,lin2021mcunetv2,wu2019fbnet,wan2020fbnetv2,Wang_2020_CVPR}. 
Next, we include depth/operators in the search space, considering two alternatives to the inverted bottleneck: 1) regular $3$$\times$$3$ convolution, 2) identity.
We target a 0.5MB model. \texttt{UDC} chooses inverted bottleneck blocks everywhere, such that the model found by including depth/operators in the search is identical to the one found when they are excluded. 
We compare \texttt{UDC} with non-uniform quantization approaches (Appendix \ref{section:non uniform quantization}), and find that
%The results indicate that 
\texttt{UDC} models
% generates 
are pareto-dominant even though approaches like HAQ \cite{wang2018haq} and \cite{yang2020automatic} employ a more expressive, non-uniform quantization which cannot be deployed on commodity MCUs and NPUs with integer math operations. 
Finally, comparing \texttt{UDC} to a SOTA unstructured pruning algorithm (Appendix \ref{section:unstructured pruning}) shows that \texttt{UDC} finds considerably more accurate models.

\textbf{SR}
The purpose of the SR experiment is to: 1) show that \texttt{UDC} can be applied to regression problems, 2) demonstrate \texttt{UDC} in a setup which constrains the computational complexity of the NN, measured in number of multiply-and-accumulate (MAC) operations, instead of model size. 
Note that $1$ MAC $= 2$ floating point operations (FLOPs), but authors often conflate the two terms. 
This experiment excludes bitwidth and sparsity from the search space, which do not impact the inference compute cost. The topology of our search space is inspired by FSRCNN \cite{dong2016accelerating}, with search over depth, width, and kernel size. 
We do all search/training on div2k and report results on set14 (see Appendix \ref{section:Super Resolution experiment settings} for results on div2k and set5). 
We target 4x upscaling and report MACs for an input patch of 64x64. 
We train all NNs using the same settings. 
Fig. \ref{fig:super resolution} shows that \texttt{UDC} finds Pareto-dominant NNs compared with SOTA efficient SR methods SESR, FSRCNN, and TPSR \cite{bhardwaj2021collapsible,dong2016accelerating,lee2020journey}.
%\vspace{-1em}

%\vspace{-0.3em}
\subsection{Comparison with random search}
\label{subsection:comparison with random search}
%\vspace{-1em}
Random search is a strong baseline that can compete with DNAS in some settings \cite{li2020random}.
We are not able to add early stopping \cite{li2020random}, since sparsity is only applied during stages 2-3, when it is too late to save time from stopping. 
Fig. \ref{fig:random baseline scatter} compares \texttt{UDC} to randomly generated NNs constrained to model sizes $<0.55$MB for CIFAR100, with the search space used in Fig. \ref{fig:CIFAR100 results}. 
The gap between the best \texttt{UDC} models and random search is $4.77\%$.
The best accuracy achieved by random search after a given number of trials is shown in Fig \ref{fig:random baseline bootstrap}.
To integrate out randomness from the the ordering of trials, we permute the order of the random search results $100\times$, averaging over the permutations. 
Since \texttt{UDC} search requires $\sim$$2$$\times$ longer to find a NN than training a baseline NN, we plot the cost of \texttt{UDC} as 3 trials (2 to search and 1 to finetune). These results confirm that \texttt{UDC} finds better NNs faster than random search.

DNAS alternatives like evolutionary search require training many NNs or proxies to relax the compute burden \cite{elsken2019neural,elsken2018efficient,cai2019once,lin2020mcunet,lin2021mcunetv2}. 
However, using proxies assumes a search space dense with performant NNs, which is not true in our case (Fig. \ref{fig:random baseline scatter}).% shows that small, accurate NNs are rare in our search space, which motivates our use of DNAS.
%\vspace{-1em}
\subsection{Validation of algorithm and system components}
\label{section:ablations}

%\vspace{-0.4em}
\begin{table}
\centering
\caption{
Ablation on \texttt{UDC} components for experiment targeting $0.5$MB models. 
$*$--Constraint not met, so accuracy not reported. 
$\dagger$--GPUD not reported as hyperparameter search for $\lambda$ is difficult to account for. 
For ImageNet, GPUD is normalized to the typical training cost of MBNetV2 ($460$ images / s) \cite{mbnetv2_training_speed}) 
to allow for comparison to other works. 
Dashes indicate the experiment is prohibitively expensive to run because it requires hyperparameter search for $\lambda$.
}
% \vspace{0.5em}
\label{table:ablation}
\resizebox{\linewidth}{!}{%
\begin{tabular}{lcccccccc}
\toprule
%& \multicolumn{4}{c}{--------------------------------------  Features  --------------------------------------}  & \multicolumn{2}{c}{CIFAR100} & \multicolumn{2}{c}{ImageNet} \\
%& Rejection Sampling    & Multiple MC Samples   & $\pi$ Projection    & No $\lambda$ Tuning & Acc. (\%) & GPUD & Acc. (\%) & GPUD\\& 
& \textbf{Rejection} & \textbf{Multiple}     & \textbf{$\pi$--}     & \textbf{No $\lambda$} & \multicolumn{2}{c}{\textbf{CIFAR100}} & \multicolumn{2}{c}{\textbf{ImageNet}} \\
& \textbf{Sampling}  & \textbf{MC Samples}   & \textbf{Projection}  & \textbf{Tuning}       & \textbf{Acc. (\%)} & \textbf{GPUD} & \textbf{Acc. (\%)} & \textbf{GPUD} \\
\midrule
Vanilla DNAS & & & & & * & $\dagger$ & --- & --- \\
 \midrule
Partial \texttt{UDC} & \checkmark & \checkmark & & \checkmark   & * & $\dagger$ & * & 3.2 \\
Partial \texttt{UDC} & & \checkmark & \checkmark & \checkmark   & {79.14} & 1.17 & * & 3.2 \\
Partial \texttt{UDC} & & & \checkmark & \checkmark   & {79.06} & 6.21 & --- & --- \\
Partial \texttt{UDC} & & \checkmark & & \checkmark  & * & 
%\cellcolor{blue!25}
\textbf{0.83} & * & 3.2 \\
Partial \texttt{UDC} & \checkmark & \checkmark & & & {76.48} & $\dagger$ & --- & --- \\
Partial \texttt{UDC} & & \checkmark & & & * & $\dagger$ & --- & --- \\ \midrule
\textbf{Full \texttt{UDC}} & \checkmark & \checkmark & \checkmark & \checkmark   
& %\cellcolor{blue!25}\
\textbf{79.24} & 1.17 & 
%\cellcolor{blue!25} 
\textbf{{64.09}} & 
%\cellcolor{blue!25} 
\textbf{3.2} \\
\bottomrule
\end{tabular}}
% \vspace{-1em}
\end{table}
\begin{wraptable}{r}{0.6\textwidth}
% \vspace{-1em}
\caption{Compressed model size (MB). Vela size represents Flash usage when deployed to NPU. MBNetV2 listed for reference.}
\label{table:vela}
\centering
\resizebox{0.6\textwidth}{!}{%
\begin{tabular}{lcccc}
\toprule
        & \textbf{Original} & \textbf{RHS of \eqref{eq:practical entropy}}  & \textbf{RHS of \eqref{eq:practical entropy} \&} & \textbf{Vela (Ratio)} \\  
        & & & \textbf{Compressed $\bm{m}$} & \\ 
        \midrule
Model 1 & 2.32  & 0.49 & 0.49 & 0.44 (5.27$\times$) \\
Model 2 & 2.22 & 0.9 & 0.9 &  0.83 (2.67$\times$) \\
Model 3 & 5.23 & 1.27 & 1.27 & 1.15 (4.55$\times$) \\
MBNetV2 & 3.31 & 3.31 & 3.31 & 2.87 (1.15$\times$) \\
\bottomrule
\end{tabular}
}
%\vspace{-1em}
\end{wraptable}

\textbf{Compression}
We verified our model size approximation, i.e. the RHS of \eqref{eq:practical entropy} with pruning mask compressed to the entropy limit, by 
%All model sizes are computed using the RHS of \eqref{eq:practical entropy} assuming that the pruning mask can be compressed to the entropy limit. 
%To verify this, we 
compressing $\bm{m}$ using arithmetic coding \cite{mentzer2019practical} for all \texttt{UDC} ImageNet models (Fig. \ref{fig:imagenet}).
We observed that theoretical and practical compressed model sizes are within two decimal places (Table \ref{table:vela}).
%\textbf{Measurements on NPU} 
Table \ref{table:vela} shows that Vela can achieve compression ratios even larger than those predicted by \texttt{UDC} for NPU deployment, which stems from the fact that non-zero weights can often be compressed to less than $b$ bits per weight by Golomb-Rice compression.

\textbf{Ablations} 
%To measure the impact of the individual components in Section \ref{section:dnas}, 
Table \ref{table:ablation} gives an ablation on \texttt{UDC} components (Sec.~\ref{section:dnas}) on CIFAR100 and ImageNet,
%experiments ().
%Each ablation variant was 
which we evaluate on final accuracy, 
%In addition to final task accuracy, we evaluate methods based on whether or not they require a 
hyperparameter search over $\lambda$ requirement, and runtime in GPU days (GPUD) normalized to a typical training setup. 
Table \ref{table:ablation} shows that disabling any of the \texttt{UDC} components leads to significantly worse results on both
datasets.
%CIFAR100 and ImageNet. 
%The first row represents 
Vanilla DNAS \cite{dong2019network} fails to meet the HW constraint.
%on both 
% datasets.
%CIFAR100 and ImageNet. 
Fig. \ref{fig:number format} shows the isolated benefit of the proposed number format over the baseline \eqref{eq:bitwidth search}.

\textbf{Runtime} An important aspect in evaluating NAS algorithms is their runtime, or the time required to yield results. Part of the challenge in comparing algorithm runtimes is that the runtime generally depends on: 1) software implementation quality and HW platform, which jointly form the “system” that the algorithm runs on, 2) the algorithm itself, i.e. what the algorithm is actually doing during the search, 3) the number of epochs (or amount of data processed) for which the search is run. The challenge is that only 2-3 are algorithm dependent, whereas 1) depends on the code quality and the resources of the experimenter (i.e. higher grade GPUs exhibit higher throughput compared to low-end GPUs).

% In absolute terms In the paper, we tried to address the runtime question by saying that UDC runs at roughly half the speed (230 images / s) of a typical training experiment (460 images / s) on ImageNet. Although we did not state this in the paper, we also observed that UDC’s speed is nearly independent of the search space size in the ImageNet experiment. In other words, introducing quantization and unstructured pruning into the search does not lead to a slowdown compared to only searching for width. Likewise, increasing the number of options (i.e. how many distinct sparsity rates to consider) also does not lead to a slowdown. This suggests that UDC’s search speed is highly scalable with the search space size and that most of the slowdown can be accounted for by the system which the algorithm is running on. For that reason, we list the search cost of UDC in the last column of Table 3 as 3.2 GPUD normalized to a system running at 460 images / s.

In order to compare with other works, we now attempt to disentangle the 3 components that make up the runtime in Table \ref{table:runtime}. We list the system and algorithm specific search speed measured in images per second, when available from the reference, the number of search epochs, the search cost in GPUD under the system used in the reference, and the search cost in GPUD under a common system assumed to be running at $460$ images / s \cite{mbnetv2_training_speed}. For several references, the algorithm begins with a pretrained model, which we assume is trained for a standard $200$ epochs \cite{mbnetv2_training_speed}. Table \ref{table:runtime} shows that in absolute terms (i.e. the system specific cost), \texttt{UDC} is $1.4 \times$ faster than FBNet and FBNetV2. When comparing approaches based on a normalized system running at $460$ images / s, \texttt{UDC} is faster than all of the competing approaches other than FBNet and FBNetV2. Our hope is that Table \ref{table:runtime} gives a rough sense of the relative search cost of \texttt{UDC} and the competing methods.

\begin{table}
\caption{Algorithm runtime comparison on ImageNet. System is defined as software implementation and HW platform. Algorithms which begin with a pretrained model are assumed to pretrain the model for $200$ epochs \cite{mbnetv2_training_speed} and are marked by $\dagger$.}
\label{table:runtime}
\resizebox{\linewidth}{!}{%
\begin{tabular}{ccccc}
\toprule \\
& \begin{tabular}[c]{@{}l@{}} \textbf{Images / s} \\ \textbf{(system \& alg. specific)}\end{tabular} & \begin{tabular}[c]{@{}l@{}}\textbf{Search epochs} \\ \textbf{(alg. specific)}\end{tabular} & \begin{tabular}[c]{@{}l@{}}\textbf{GPUD} \\ \textbf{(system and alg. specific)}\end{tabular} & \begin{tabular}[c]{@{}l@{}}\textbf{GPUD normalized to} \\ \textbf{460 im. / s (alg. specific)}\end{tabular} \\ \hline
\texttt{UDC} & 230 & 100 & 6.4 & 3.2 \\
FBNet \cite{wu2019fbnet} & 14.8 (11.5e6 images over 216 hours) & 90 (on 1/10 of ImageNet classes)& 9 & 0.3\\
FBNetV2 \cite{wan2020fbnetv2} & 14.8 (11.5e6 images over 216 hours) & 90 (on 1/10 of ImageNet classes) & 9 & 0.3 \\
MCUNet \cite{lin2020mcunet} & --- & 450 & --- & 14.5 \\
MCUNetV2 \cite{lin2021mcunetv2} & --- & 450 & --- & 14.5 \\
Choi et al. \cite{choi2020learning} & --- & $212.5^\dagger$ & --- & 6.8 \\
Uhlich et al. \cite{Uhlich2020Mixed} & --- & $250^\dagger$ & --- & 8 \\
\bottomrule
\end{tabular}}
\end{table}

\textbf{Societal impact \& limitations} 
Developing \texttt{UDC} used hundreds of energy-consuming GPU hours. However, this can be amortized by increasing the energy efficiency of billions of IoT devices. 
A limitation of \texttt{UDC} is that we retrain $\braces{\bm{\theta}}$ for every constraint $e^*$, whereas approaches based on OFA can amortize the cost of training $\braces{\bm{\theta}}$ across multiple constraints.
%\vspace{-1.5em}
\section{Conclusion}
%\vspace{-1.5em}
Emerging NPU HW platforms specialized for TinyML support model compression, whereby quantized and pruned NNs can be stored in a reduced memory footprint.
While compression is highly desirable, it increases the complexity of the NN design process, as the space of candidate NNs is increased by adding quantization and pruning on top of the conventional NN architecture choices.
To enable TinyML practitioners to fully exploit HW model compression in NPUs, we describe a unified DNAS framework to search both architecture choices and aggressive per-layer quantization and pruning. 
We describe a number of improvements on top of DNAS, allowing us to demonstrate SOTA TinyML models that fully exploit model compression, as well as a comparison with random sampling and extensive ablations.

\newpage

% \section*{References}

\bibliographystyle{plainnat}
\bibliography{egbib}

\begin{thebibliography}{72}
\providecommand{\natexlab}[1]{#1}
\providecommand{\url}[1]{\texttt{#1}}
\expandafter\ifx\csname urlstyle\endcsname\relax
  \providecommand{\doi}[1]{doi: #1}\else
  \providecommand{\doi}{doi: \begingroup \urlstyle{rm}\Url}\fi

\bibitem[ali()]{alif}
{ Alif Semiconductor: Introducing the Ensemble and Crescendo families of fusion
  processors and microcontrollers }.
\newblock \url{ https://alifsemi.com/products/ }.
\newblock Accessed: 2019-05-02.

\bibitem[eth({\natexlab{a}})]{ethos-u55}
{Arm Ethos-U55 Micro Neural Processing Unit (uNPU)}.
\newblock \url{ https://www.arm.com/products/silicon-ip-cpu/ethos/ethos-u55 },
  {\natexlab{a}}.
\newblock Accessed: 2021-10-26.

\bibitem[eth({\natexlab{b}})]{ethos-vela}
{Arm Vela Tool for Ethos-U Micro Neural Processing Unit (uNPU)}.
\newblock \url{ https://pypi.org/project/ethos-u-vela/ }, {\natexlab{b}}.
\newblock Accessed: 2021-10-26.

\bibitem[mbn()]{mbnetv2_training_speed}
{Reproduction of MobileNetV2 using MXNet }.
\newblock \url{ https://github.com/liangfu/mxnet-mobilenet-v2 }.
\newblock Accessed: 2022-10-2.

\bibitem[pra()]{practical2020tinyml}
{Practical application of tinyML in battery powered anomaly sensors for
  predictive maintenance of industrial assets}.
\newblock
  \url{https://cms.tinyml.org/wp-content/uploads/talks2020/tinyML_Talks_Mark_Stubbs_200818.pdf}.
\newblock Accessed: 2021-10-09.

\bibitem[stm()]{stm32}
{STM32 Hardware Specification, Wikipedia}.
\newblock \url{https://en.wikipedia.org/wiki/STM32}.
\newblock Accessed: 2019-05-02.

\bibitem[tfl()]{tflitemicro}
{TensorFlow Lite for Microcontrollers}.
\newblock
  \url{https://github.com/tensorflow/tensorflow/tree/master/tensorflow/lite/experimental/micro}.
\newblock Accessed: 2019-05-02.

\bibitem[ute()]{utensor}
{uTensor}.
\newblock \url{http://utensor.ai/}.
\newblock Accessed: 2019-05-02.

\bibitem[vis()]{visual_wake}
{Visual Wake Words Challenge, CVPR 2019}.
\newblock URL
  \url{https://docs.google.com/document/u/2/d/e/2PACX-1vStp3uPhxJB0YTwL4T__Q5xjclmrj6KRs55xtMJrCyi82GoyHDp2X0KdhoYcyjEzKe4v75WBqPObdkP/pub}.
\newblock Accessed: 2019-05-02.

\bibitem[Agustsson and Timofte(2017)]{agustsson2017ntire}
Eirikur Agustsson and Radu Timofte.
\newblock Ntire 2017 challenge on single image super-resolution: Dataset and
  study.
\newblock In \emph{IEEE/CVF CVPR workshops}, pages 126--135, 2017.

\bibitem[Ahmad et~al.(2020)Ahmad, Arif, Hanif, Hafiz, and
  Shafique]{ahmad2020superslash}
Hazoor Ahmad, Tabasher Arif, Muhammad~Abdullah Hanif, Rehan Hafiz, and Muhammad
  Shafique.
\newblock Superslash: A unified design space exploration and model compression
  methodology for design of deep learning accelerators with reduced off-chip
  memory access volume.
\newblock \emph{IEEE Transactions on Computer-Aided Design of Integrated
  Circuits and Systems}, 39\penalty0 (11):\penalty0 4191--4204, 2020.

\bibitem[Arora et~al.(2019)Arora, Li, and Lyu]{arora2018theoretical}
Sanjeev Arora, Zhiyuan Li, and Kaifeng Lyu.
\newblock Theoretical analysis of auto rate-tuning by batch normalization.
\newblock In \emph{International Conference on Learning Representations}, 2019.
\newblock URL \url{https://openreview.net/forum?id=rkxQ-nA9FX}.

\bibitem[Banbury et~al.(2021{\natexlab{a}})Banbury, Reddi, Torelli, Jeffries,
  Kiraly, Holleman, Montino, Kanter, Warden, Pau, Thakker, antonio torrini, jay
  cordaro, Guglielmo, Duarte, Tran, Tran, niu wenxu, and
  xu~xuesong]{banbury2021mlperf}
Colby Banbury, Vijay~Janapa Reddi, Peter Torelli, Nat Jeffries, Csaba Kiraly,
  Jeremy Holleman, Pietro Montino, David Kanter, Pete Warden, Danilo Pau,
  Urmish Thakker, antonio torrini, jay cordaro, Giuseppe~Di Guglielmo, Javier
  Duarte, Honson Tran, Nhan Tran, niu wenxu, and xu~xuesong.
\newblock {MLP}erf tiny benchmark.
\newblock In \emph{Thirty-fifth Conference on Neural Information Processing
  Systems Datasets and Benchmarks Track (Round 1)}, 2021{\natexlab{a}}.
\newblock URL \url{https://openreview.net/forum?id=8RxxwAut1BI}.

\bibitem[Banbury et~al.(2021{\natexlab{b}})Banbury, Zhou, Fedorov, Matas,
  Thakker, Gope, Janapa~Reddi, Mattina, and Whatmough]{banbury2021micronets}
Colby Banbury, Chuteng Zhou, Igor Fedorov, Ramon Matas, Urmish Thakker, Dibakar
  Gope, Vijay Janapa~Reddi, Matthew Mattina, and Paul Whatmough.
\newblock {MicroNets: Neural network architectures for deploying tinyml
  applications on commodity microcontrollers}.
\newblock \emph{Proceedings of Machine Learning and Systems}, 3,
  2021{\natexlab{b}}.

\bibitem[Bhardwaj et~al.(2022)Bhardwaj, O'Neil, Gope, Matas, Chalfin, Suda,
  Meng, Loh, and Milosavljevic]{bhardwaj2021collapsible}
Kartikeya Bhardwaj, Liam O'Neil, Dibakar Gope, Ramon Matas, Alex Chalfin,
  Naveen Suda, Lingchuan Meng, Danny Loh, and Milos Milosavljevic.
\newblock Collapsible linear blocks for super-efficient super resolution.
\newblock \emph{Proceedings of Machine Learning and Systems}, 4, 2022.

\bibitem[Blalock et~al.(2020)Blalock, Gonzalez~Ortiz, Frankle, and
  Guttag]{MLSYS2020_d2ddea18}
Davis Blalock, Jose~Javier Gonzalez~Ortiz, Jonathan Frankle, and John Guttag.
\newblock What is the state of neural network pruning?
\newblock In I.~Dhillon, D.~Papailiopoulos, and V.~Sze, editors,
  \emph{Proceedings of Machine Learning and Systems}, volume~2, pages 129--146,
  2020.
\newblock URL
  \url{https://proceedings.mlsys.org/paper/2020/file/d2ddea18f00665ce8623e36bd4e3c7c5-Paper.pdf}.

\bibitem[Cai et~al.(2020)Cai, Gan, Wang, Zhang, and Han]{cai2019once}
Han Cai, Chuang Gan, Tianzhe Wang, Zhekai Zhang, and Song Han.
\newblock Once-for-all: Train one network and specialize it for efficient
  deployment.
\newblock In \emph{International Conference on Learning Representations}, 2020.
\newblock URL \url{https://openreview.net/forum?id=HylxE1HKwS}.

\bibitem[Choi et~al.(2020)Choi, El-Khamy, and Lee]{choi2020learning}
Yoojin Choi, Mostafa El-Khamy, and Jungwon Lee.
\newblock Learning sparse low-precision neural networks with learnable
  regularization.
\newblock \emph{IEEE Access}, 2020.

\bibitem[Cover(1999)]{cover1999elements}
Thomas~M Cover.
\newblock \emph{Elements of information theory}.
\newblock John Wiley \& Sons, 1999.

\bibitem[Dong et~al.(2016)Dong, Loy, and Tang]{dong2016accelerating}
Chao Dong, Chen~Change Loy, and Xiaoou Tang.
\newblock Accelerating the super-resolution convolutional neural network.
\newblock In \emph{European conference on computer vision}, pages 391--407.
  Springer, 2016.

\bibitem[Dong and Yang(2019{\natexlab{a}})]{dong2019network}
Xuanyi Dong and Yi~Yang.
\newblock Network pruning via transformable architecture search.
\newblock In \emph{Advances in Neural Information Processing Systems}, pages
  759--770, 2019{\natexlab{a}}.

\bibitem[Dong and Yang(2019{\natexlab{b}})]{dong2019searching}
Xuanyi Dong and Yi~Yang.
\newblock Searching for a robust neural architecture in four gpu hours.
\newblock In \emph{IEEE/CVF CVPR}, pages 1761--1770, 2019{\natexlab{b}}.

\bibitem[Elsken et~al.(2019{\natexlab{a}})Elsken, Metzen, and
  Hutter]{elsken2018efficient}
Thomas Elsken, Jan~Hendrik Metzen, and Frank Hutter.
\newblock Efficient multi-objective neural architecture search via lamarckian
  evolution.
\newblock In \emph{International Conference on Learning Representations},
  2019{\natexlab{a}}.

\bibitem[Elsken et~al.(2019{\natexlab{b}})Elsken, Metzen, and
  Hutter]{elsken2019neural}
Thomas Elsken, Jan~Hendrik Metzen, and Frank Hutter.
\newblock Neural architecture search: A survey.
\newblock \emph{Journal of Machine Learning Research}, 20\penalty0
  (55):\penalty0 1--21, 2019{\natexlab{b}}.

\bibitem[Fedorov et~al.(2019)Fedorov, Adams, Mattina, and
  Whatmough]{fedorov2019sparse}
Igor Fedorov, Ryan~P Adams, Matthew Mattina, and Paul~N Whatmough.
\newblock {SpArSe: Sparse architecture search for CNNs on resource-constrained
  microcontrollers}.
\newblock \emph{Advances in Neural Information Processing Systems}, 32, 2019.

\bibitem[Fedorov et~al.(2020)Fedorov, Stamenovic, Jensen, Yang, Mandell, Gan,
  Mattina, and Whatmough]{fedorov2020tinylstms}
Igor Fedorov, Marko Stamenovic, Carl Jensen, Li-Chia Yang, Ari Mandell, Yiming
  Gan, Matthew Mattina, and Paul~N Whatmough.
\newblock {TinyLSTMs: Efficient Neural Speech Enhancement for Hearing Aids}.
\newblock \emph{INTERSPEECH}, 2020.

\bibitem[Gong et~al.(2019)Gong, Jiang, Wang, Lin, Liu, and Pan]{gong2019mixed}
Chengyue Gong, Zixuan Jiang, Dilin Wang, Yibo Lin, Qiang Liu, and David~Z Pan.
\newblock Mixed precision neural architecture search for energy efficient deep
  learning.
\newblock In \emph{2019 IEEE/ACM International Conference on Computer-Aided
  Design (ICCAD)}, pages 1--7. IEEE, 2019.

\bibitem[Guo et~al.(2016)Guo, Yao, and Chen]{guo2016dynamic}
Yiwen Guo, Anbang Yao, and Yurong Chen.
\newblock Dynamic network surgery for efficient dnns.
\newblock In \emph{Proceedings of the 30th International Conference on Neural
  Information Processing Systems}, pages 1387--1395, 2016.

\bibitem[Guo et~al.(2020)Guo, Zhang, Mu, Heng, Liu, Wei, and
  Sun]{guo2020single}
Zichao Guo, Xiangyu Zhang, Haoyuan Mu, Wen Heng, Zechun Liu, Yichen Wei, and
  Jian Sun.
\newblock Single path one-shot neural architecture search with uniform
  sampling.
\newblock In \emph{European Conference on Computer Vision}, pages 544--560.
  Springer, 2020.

\bibitem[Gupta et~al.(2017)Gupta, Suggala, Goyal, Simhadri, Paranjape, Kumar,
  Goyal, Udupa, Varma, and Jain]{gupta2017protonn}
Chirag Gupta, Arun~Sai Suggala, Ankit Goyal, Harsha~Vardhan Simhadri, Bhargavi
  Paranjape, Ashish Kumar, Saurabh Goyal, Raghavendra Udupa, Manik Varma, and
  Prateek Jain.
\newblock Protonn: Compressed and accurate knn for resource-scarce devices.
\newblock In \emph{Proceedings of the 34th International Conference on Machine
  Learning-Volume 70}, pages 1331--1340. JMLR. org, 2017.

\bibitem[Han et~al.(2015)Han, Pool, Tran, and Dally]{han2015learning}
Song Han, Jeff Pool, John Tran, and William Dally.
\newblock Learning both weights and connections for efficient neural network.
\newblock \emph{Advances in Neural Information Processing Systems}, 28, 2015.

\bibitem[Han et~al.(2016)Han, Mao, and Dally]{han2015deep}
Song Han, Huizi Mao, and William~J Dally.
\newblock Deep compression: Compressing deep neural networks with pruning,
  trained quantization and huffman coding.
\newblock \emph{International Conference on Learning Representations (ICLR)},
  2016.

\bibitem[He et~al.(2017)He, Zhang, and Sun]{he2017channel}
Yihui He, Xiangyu Zhang, and Jian Sun.
\newblock Channel pruning for accelerating very deep neural networks.
\newblock In \emph{Proceedings of the IEEE International Conference on Computer
  Vision}, pages 1389--1397, 2017.

\bibitem[Hong et~al.(2020)Hong, Li, Zhang, Tang, Wang, Li, and
  Yu]{hong2020dropnas}
Weijun Hong, Guilin Li, Weinan Zhang, Ruiming Tang, Yunhe Wang, Zhenguo Li, and
  Yong Yu.
\newblock Dropnas: Grouped operation dropout for differentiable architecture
  search.
\newblock In \emph{IJCAI}, pages 2326--2332, 2020.

\bibitem[Howard et~al.(2017)Howard, Zhu, Chen, Kalenichenko, Wang, Weyand,
  Andreetto, and Adam]{howard2017mobilenets}
Andrew~G. Howard, Menglong Zhu, Bo~Chen, Dmitry Kalenichenko, Weijun Wang,
  Tobias Weyand, Marco Andreetto, and Hartwig Adam.
\newblock Mobilenets: Efficient convolutional neural networks for mobile vision
  applications.
\newblock \emph{CoRR}, abs/1704.04861, 2017.
\newblock URL \url{http://arxiv.org/abs/1704.04861}.

\bibitem[Hu et~al.(2020)Hu, Xie, Zheng, Liu, Shi, Liu, and Lin]{hu2020dsnas}
Shoukang Hu, Sirui Xie, Hehui Zheng, Chunxiao Liu, Jianping Shi, Xunying Liu,
  and Dahua Lin.
\newblock Dsnas: Direct neural architecture search without parameter
  retraining.
\newblock In \emph{IEEE/CVF CVPR}, pages 12084--12092, 2020.

\bibitem[Huang et~al.(2015)Huang, Singh, and Ahuja]{Huang-CVPR-2015}
Jia-Bin Huang, Abhishek Singh, and Narendra Ahuja.
\newblock Single image super-resolution from transformed self-exemplars.
\newblock In \emph{Proceedings of the IEEE Conference on Computer Vision and
  Pattern Recognition}, pages 5197--5206, 2015.

\bibitem[Idelbayev and Carreira-Perpin{\'a}n(2020)]{idelbayev2020low}
Yerlan Idelbayev and Miguel~A Carreira-Perpin{\'a}n.
\newblock Low-rank compression of neural nets: Learning the rank of each layer.
\newblock In \emph{Proceedings of the IEEE/CVF Conference on Computer Vision
  and Pattern Recognition}, pages 8049--8059, 2020.

\bibitem[Jang et~al.(2017)Jang, Gu, and Poole]{jang2016categorical}
Eric Jang, Shixiang Gu, and Ben Poole.
\newblock Categorical reparameterization with gumbel-softmax.
\newblock \emph{{International Conference on Learning Representations}}, 2017.

\bibitem[Jin et~al.(2019)Jin, Wang, Slocum, Yang, Dai, Yan, and
  Feng]{jin2019rc}
Xiaojie Jin, Jiang Wang, Joshua Slocum, Ming-Hsuan Yang, Shengyang Dai,
  Shuicheng Yan, and Jiashi Feng.
\newblock Rc-darts: Resource constrained differentiable architecture search.
\newblock \emph{arXiv preprint arXiv:1912.12814}, 2019.

\bibitem[Krizhevsky et~al.(2009)Krizhevsky, Hinton,
  et~al.]{krizhevsky2009learning}
Alex Krizhevsky, Geoffrey Hinton, et~al.
\newblock Learning multiple layers of features from tiny images.
\newblock 2009.

\bibitem[Kumar et~al.(2017)Kumar, Goyal, and Varma]{kumar2017resource}
Ashish Kumar, Saurabh Goyal, and Manik Varma.
\newblock Resource-efficient machine learning in 2 kb ram for the internet of
  things.
\newblock In \emph{Proceedings of the 34th International Conference on Machine
  Learning-Volume 70}, pages 1935--1944. JMLR. org, 2017.

\bibitem[Kusupati et~al.(2020)Kusupati, Ramanujan, Somani, Wortsman, Jain,
  Kakade, and Farhadi]{kusupati2020soft}
Aditya Kusupati, Vivek Ramanujan, Raghav Somani, Mitchell Wortsman, Prateek
  Jain, Sham Kakade, and Ali Farhadi.
\newblock Soft threshold weight reparameterization for learnable sparsity.
\newblock In \emph{International Conference on Machine Learning}, pages
  5544--5555. PMLR, 2020.

\bibitem[Lee et~al.(2020)Lee, Dudziak, Abdelfattah, Venieris, Kim, Wen, and
  Lane]{lee2020journey}
Royson Lee, {\L}ukasz Dudziak, Mohamed Abdelfattah, Stylianos~I Venieris, Hyeji
  Kim, Hongkai Wen, and Nicholas~D Lane.
\newblock Journey towards tiny perceptual super-resolution.
\newblock In \emph{European Conference on Computer Vision}, pages 85--102.
  Springer, 2020.

\bibitem[Li et~al.(2019)Li, Bhargav, Whatmough, and Wong]{li2019chip}
Haitong Li, Mudit Bhargav, Paul~N Whatmough, and H-S~Philip Wong.
\newblock On-chip memory technology design space explorations for mobile deep
  neural network accelerators.
\newblock In \emph{2019 56th ACM/IEEE design automation conference (DAC)},
  pages 1--6. IEEE, 2019.

\bibitem[Li and Talwalkar(2020)]{li2020random}
Liam Li and Ameet Talwalkar.
\newblock Random search and reproducibility for neural architecture search.
\newblock In \emph{Uncertainty in artificial intelligence}, pages 367--377.
  PMLR, 2020.

\bibitem[Liberis et~al.(2021)Liberis, Dudziak, and Lane]{liberis2021munas}
Edgar Liberis, {\L}ukasz Dudziak, and Nicholas~D Lane.
\newblock $\mu$nas: Constrained neural architecture search for
  microcontrollers.
\newblock In \emph{Proceedings of the 1st Workshop on Machine Learning and
  Systems}, pages 70--79, 2021.

\bibitem[Lin et~al.(2020{\natexlab{a}})Lin, Chen, Lin, Gan, Han,
  et~al.]{lin2020mcunet}
Ji~Lin, Wei-Ming Chen, Yujun Lin, Chuang Gan, Song Han, et~al.
\newblock Mcunet: Tiny deep learning on iot devices.
\newblock \emph{Advances in Neural Information Processing Systems},
  33:\penalty0 11711--11722, 2020{\natexlab{a}}.

\bibitem[Lin et~al.(2021)Lin, Chen, Cai, Gan, and Han]{lin2021mcunetv2}
Ji~Lin, Wei-Ming Chen, Han Cai, Chuang Gan, and Song Han.
\newblock Mcunetv2: Memory-efficient patch-based inference for tiny deep
  learning.
\newblock \emph{arXiv preprint arXiv:2110.15352}, 2021.

\bibitem[Lin et~al.(2020{\natexlab{b}})Lin, Stich, Barba, Dmitriev, and
  Jaggi]{Lin2020Dynamic}
Tao Lin, Sebastian~U. Stich, Luis Barba, Daniil Dmitriev, and Martin Jaggi.
\newblock Dynamic model pruning with feedback.
\newblock In \emph{International Conference on Learning Representations},
  2020{\natexlab{b}}.
\newblock URL \url{https://openreview.net/forum?id=SJem8lSFwB}.

\bibitem[Liu et~al.(2019)Liu, Simonyan, and Yang]{liu2018darts}
Hanxiao Liu, Karen Simonyan, and Yiming Yang.
\newblock {DARTS}: Differentiable architecture search.
\newblock In \emph{ICLR}, 2019.

\bibitem[McDonnell(2018)]{mcdonnell2018training}
Mark~D. McDonnell.
\newblock Training wide residual networks for deployment using a single bit for
  each weight.
\newblock In \emph{International Conference on Learning Representations}, 2018.
\newblock URL \url{https://openreview.net/forum?id=rytNfI1AZ}.

\bibitem[Mentzer et~al.(2019)Mentzer, Agustsson, Tschannen, Timofte, and
  Van~Gool]{mentzer2019practical}
Fabian Mentzer, Eirikur Agustsson, Michael Tschannen, Radu Timofte, and Luc
  Van~Gool.
\newblock Practical full resolution learned lossless image compression.
\newblock In \emph{IEEE/CVF CVPR}, 2019.

\bibitem[Molchanov et~al.(2016)Molchanov, Tyree, Karras, Aila, and
  Kautz]{molchanov2016pruning}
Pavlo Molchanov, Stephen Tyree, Tero Karras, Timo Aila, and Jan Kautz.
\newblock Pruning convolutional neural networks for resource efficient
  inference.
\newblock \emph{arXiv preprint arXiv:1611.06440}, 2016.

\bibitem[Paulus et~al.(2020)Paulus, Maddison, and Krause]{paulus2020rao}
Max~B Paulus, Chris~J Maddison, and Andreas Krause.
\newblock Rao-blackwellizing the straight-through gumbel-softmax gradient
  estimator.
\newblock \emph{arXiv preprint arXiv:2010.04838}, 2020.

\bibitem[Pham et~al.(2018)Pham, Guan, Zoph, Le, and Dean]{pham2018efficient}
Hieu Pham, Melody Guan, Barret Zoph, Quoc Le, and Jeff Dean.
\newblock Efficient neural architecture search via parameters sharing.
\newblock In \emph{International Conference on Machine Learning}, pages
  4095--4104. PMLR, 2018.

\bibitem[Russakovsky et~al.(2015)Russakovsky, Deng, Su, Krause, Satheesh, Ma,
  Huang, Karpathy, Khosla, Bernstein, Berg, and Fei-Fei]{ILSVRC15}
Olga Russakovsky, Jia Deng, Hao Su, Jonathan Krause, Sanjeev Satheesh, Sean Ma,
  Zhiheng Huang, Andrej Karpathy, Aditya Khosla, Michael Bernstein,
  Alexander~C. Berg, and Li~Fei-Fei.
\newblock {ImageNet Large Scale Visual Recognition Challenge}.
\newblock \emph{International Journal of Computer Vision (IJCV)}, 115\penalty0
  (3):\penalty0 211--252, 2015.
\newblock \doi{10.1007/s11263-015-0816-y}.

\bibitem[Sandler et~al.(2018)Sandler, Howard, Zhu, Zhmoginov, and
  Chen]{sandler2018mobilenetv2}
Mark Sandler, Andrew Howard, Menglong Zhu, Andrey Zhmoginov, and Liang-Chieh
  Chen.
\newblock Mobilenetv2: Inverted residuals and linear bottlenecks.
\newblock In \emph{IEEE/CVF CVPR}, pages 4510--4520, 2018.

\bibitem[Stock et~al.(2021)Stock, Fan, Graham, Grave, Gribonval, Jegou, and
  Joulin]{stock2021training}
Pierre Stock, Angela Fan, Benjamin Graham, Edouard Grave, R{\'e}mi Gribonval,
  Herve Jegou, and Armand Joulin.
\newblock Training with quantization noise for extreme model compression.
\newblock In \emph{International Conference on Learning Representations}, 2021.
\newblock URL \url{https://openreview.net/forum?id=dV19Yyi1fS3}.

\bibitem[Theis et~al.(2017)Theis, Shi, Cunningham, and
  Husz{\'a}r]{theis2017lossy}
Lucas Theis, Wenzhe Shi, Andrew Cunningham, and Ferenc Husz{\'a}r.
\newblock Lossy image compression with compressive autoencoders.
\newblock \emph{arXiv preprint arXiv:1703.00395}, 2017.

\bibitem[Uhlich et~al.(2020)Uhlich, Mauch, Cardinaux, Yoshiyama, Garcia,
  Tiedemann, Kemp, and Nakamura]{Uhlich2020Mixed}
Stefan Uhlich, Lukas Mauch, Fabien Cardinaux, Kazuki Yoshiyama, Javier~Alonso
  Garcia, Stephen Tiedemann, Thomas Kemp, and Akira Nakamura.
\newblock Mixed precision dnns: All you need is a good parametrization.
\newblock In \emph{International Conference on Learning Representations}, 2020.
\newblock URL \url{https://openreview.net/forum?id=Hyx0slrFvH}.

\bibitem[Wan et~al.(2020)Wan, Dai, Zhang, He, Tian, Xie, Wu, Yu, Xu, Chen,
  et~al.]{wan2020fbnetv2}
Alvin Wan, Xiaoliang Dai, Peizhao Zhang, Zijian He, Yuandong Tian, Saining Xie,
  Bichen Wu, Matthew Yu, Tao Xu, Kan Chen, et~al.
\newblock Fbnetv2: Differentiable neural architecture search for spatial and
  channel dimensions.
\newblock In \emph{IEEE/CVF CVPR}, pages 12965--12974, 2020.

\bibitem[Wang et~al.(2019)Wang, Liu, Lin, Lin, and Han]{wang2018haq}
Kuan Wang, Zhijian Liu, Yujun Lin, Ji~Lin, and Song Han.
\newblock {HAQ:} hardware-aware automated quantization.
\newblock In \emph{IEEE/CVF CVPR}, 2019.

\bibitem[Wang et~al.(2020)Wang, Wang, Cai, Lin, Liu, Wang, Lin, and
  Han]{Wang_2020_CVPR}
Tianzhe Wang, Kuan Wang, Han Cai, Ji~Lin, Zhijian Liu, Hanrui Wang, Yujun Lin,
  and Song Han.
\newblock Apq: Joint search for network architecture, pruning and quantization
  policy.
\newblock In \emph{IEEE/CVF CVPR}, June 2020.

\bibitem[Wu et~al.(2019)Wu, Dai, Zhang, Wang, Sun, Wu, Tian, Vajda, Jia, and
  Keutzer]{wu2019fbnet}
Bichen Wu, Xiaoliang Dai, Peizhao Zhang, Yanghan Wang, Fei Sun, Yiming Wu,
  Yuandong Tian, Peter Vajda, Yangqing Jia, and Kurt Keutzer.
\newblock Fbnet: Hardware-aware efficient convnet design via differentiable
  neural architecture search.
\newblock In \emph{IEEE/CVF CVPR}, pages 10734--10742, 2019.

\bibitem[Yang et~al.(2020)Yang, Gui, Zhu, and Liu]{yang2020automatic}
Haichuan Yang, Shupeng Gui, Yuhao Zhu, and Ji~Liu.
\newblock Automatic neural network compression by sparsity-quantization joint
  learning: A constrained optimization-based approach.
\newblock In \emph{IEEE/CVF CVPR}, pages 2178--2188, 2020.

\bibitem[Yang et~al.(2017)Yang, Chen, and Sze]{yang2017designing}
Tien-Ju Yang, Yu-Hsin Chen, and Vivienne Sze.
\newblock Designing energy-efficient convolutional neural networks using
  energy-aware pruning.
\newblock In \emph{IEEE/CVF CVPR}, pages 5687--5695, 2017.

\bibitem[Ye et~al.(2018)Ye, Lu, Lin, and Wang]{ye2018rethinking}
Jianbo Ye, Xin Lu, Zhe Lin, and James~Z. Wang.
\newblock Rethinking the smaller-norm-less-informative assumption in channel
  pruning of convolution layers.
\newblock In \emph{International Conference on Learning Representations}, 2018.
\newblock URL \url{https://openreview.net/forum?id=HJ94fqApW}.

\bibitem[Yu et~al.(2020)Yu, Han, Li, Shi, Cheng, and Fan]{yu2020search}
Haibao Yu, Qi~Han, Jianbo Li, Jianping Shi, Guangliang Cheng, and Bin Fan.
\newblock Search what you want: Barrier panelty nas for mixed precision
  quantization.
\newblock \emph{arXiv preprint arXiv:2007.10026}, 2020.

\bibitem[Zagoruyko and Komodakis(2016)]{zagoruyko2016wide}
Sergey Zagoruyko and Nikos Komodakis.
\newblock Wide residual networks.
\newblock In Edwin R.~Hancock Richard C.~Wilson and William A.~P. Smith,
  editors, \emph{Proceedings of the British Machine Vision Conference (BMVC)},
  pages 87.1--87.12. BMVA Press, September 2016.
\newblock ISBN 1-901725-59-6.
\newblock \doi{10.5244/C.30.87}.
\newblock URL \url{https://dx.doi.org/10.5244/C.30.87}.

\bibitem[Zhou et~al.(2017)Zhou, Yao, Guo, Xu, and Chen]{zhou2017incremental}
Aojun Zhou, Anbang Yao, Yiwen Guo, Lin Xu, and Yurong Chen.
\newblock Incremental network quantization: Towards lossless cnns with
  low-precision weights.
\newblock \emph{{International Conference on Learning Representations}}, 2017.

\bibitem[Zhu and Gupta(2017)]{zhu2017prune}
Michael Zhu and Suyog Gupta.
\newblock To prune, or not to prune: exploring the efficacy of pruning for
  model compression.
\newblock \emph{arXiv preprint arXiv:1710.01878}, 2017.

\end{thebibliography}

%%%%%%%%%%%%%%%%%%%%%%%%%%%%%%%%%%%%%%%%%%%%%%%%%%%%%%%%%%%%

\appendix
\clearpage
\section{Proof of Lemma \ref{lemma:regularizer-constraint}}
\label{sec:proof lemma 1}
\begin{proof}
We know that $\func{p}{\func{\mathcal{E}}{\braces{\bm{z}}} \neq e^*}$ must be 0. But if $\func{\mathcal{E}}{\braces{\gamma(\bm{\pi})}} \neq e^*$, $\func{p}{\func{\mathcal{E}}{\braces{\bm{z}}} \neq e^*} > 0$, which is a contradiction.
\end{proof}

\section{Proof of Lemma \ref{lemma:one hot}}
\label{sec:proof lemma 2}
\begin{proof}
If there is a $\pi_j$ which is not one-hot, then the following configuration sample has non-zero probability:
\begin{itemize}
    \item $z_j^s \neq s_j^{s'}$
    \item $z_k^s = z_k^{s'} \forall k \neq j$
\end{itemize}
Since we assumed that $\func{\mathcal{L}_{\mathcal{E}}}{\braces{\bm{z}}}=0$, Lemma \ref{lemma:regularizer-constraint} gives that $\func{{\mathcal{E}}}{\braces{\bm{z}^s}}= e^* = \func{{\mathcal{E}}}{\braces{\bm{z}^{s'}}}$. But this is a contradiction since we assumed there are no decisions for which two options have the same efficiency.
\end{proof}

\section{Derivation of entropy bound}
\label{section:derivation of entropy bound}
The entropy bound in \eqref{eq:practical entropy} can be derived as follows:
\begin{align}
    &\func{H}{Q(\bm{\theta},b,r) \odot \bm{m} } \times \norm{\bm{w}}_0 \\
    &\leq \left(\func{H}{Q(\bm{\theta},b,r)} + \func{H}{\bm{m}}\right) \times \norm{\bm{w}}_0 \\
    &\leq (b + \underbrace{\func{H}{\bm{m}}}_{-s \log_2 s - (1-s)\log_2(1-s)}) \times \norm{w}_0 
\end{align}
where the first inequality follows from the fact that the entropy of a product of RVs is bounded by the sum of their entropies and the second bound follows from the fact that $Q(\bm{\theta},b,r)$ costs at most $b$ bits per element to encode.

\section{Avoiding co-adaptation in DNAS}
\label{section:avoiding co-adaptation}
While samples of $\bm{z}$ are one-hot, samples of $\hat{\bm{z}}$ are not. This property can cause issues for an approach like DNAS, where weight-sharing can lead to co-adapation between search options \cite{hong2020dropnas,guo2020single}. The result is a large performance drop when finetuning $\braces{\gamma(\bm{\bm{\pi}})}$, compared to the value of \task{} achieved by solving \eqref{eq:regularized objective} \cite{hu2020dsnas}. One solution, which we adopt, is to use a straight-through-estimator (STE), whereby 
\begin{align}\label{eq:ste}
    \hat{z}_{f_0}[k] = \begin{cases} 
      \hat{z}[k] & k \in \func{\text{topk}}{\hat{\bm{z}},\kappa} \\
      0 & \text{else} \\
   \end{cases} \; \; \; \; \; | \; \; \; \; \; \; \; \hat{\bm{z}}_f = \frac{\hat{z}_{f_0}}{ \norm{\hat{z}_{f_0}}_1 }
\end{align}
is used in the forward pass, where $\kappa$ is the number of non-zeros in $\hat{z}_f[k]$, $\text{topk}(\hat{\bm{z}},\kappa)$ returns the indices of the $\kappa$ largest elements of $\hat{\bm{z}}$, and $\hat{\bm{z}}$ is used in the backward pass \cite{jang2016categorical}. Typically, $\kappa \in \braces{1,2}$.

\section{CIFAR100 experiment settings}
\label{section:cifar100 experiment settings}
We run the search for 200 epochs, annealing $\tau$ from $0.66$ to $0.1$ using an exponential schedule. We use SGD for $\bm{\theta}$ with learning rate annealed from $0.1$ to $1e-4$ using a cosine schedule and we use ADAM for $\braces{\pi}$ with a constant learning rate of $1e-3$. We increase $\vartheta$ from $0$ to $0.5$ using a linear schedule and we increase $\zeta^t$ from $0.1$ to $1$ using a linear schedule. We initialize the search by running a warmup stage for 50 epochs where we use SGD for $\bm{\theta}$ with learning rate annealed from $0.1$ to $1e-4$ using a cosine schedule, $\tau$ is annealed from $0.66$ to $0.1$ using an exponential schedule, $\braces{\pi}$ is not learned, and $\vartheta=0$. During both warmup and search, we set $\kappa=K$ for all width decisions and $\kappa=2$ for all quantization and sparsity decisions.

To finetune the discovered models, we run stage 1 for 254 epochs using SGD and cosine decay with restarts learning rate schedule, cycling between $0.1$ and $1e-4$ at intervals which double after every cycle and beginning with a cycle of 2 epochs. We run stage 2 for 60 epochs, annealing the learning rate from $0.1$ to $1e-4$ using a cosine schedule and then we run stage 3 for 30 epochs, annealing the learning rate from $1e-4$ to $0$ using a cosine schedule. We use distillation with a teacher model whose architecture is WRN 20-10. For data augmentation, we use horizontal flipping, random crop with a size of 4, and cutout with a size of 18. We use $\ell_2$ regularization with coefficient $5e-4$. We disable learning of batchnorm scale and offset parameters \cite{mcdonnell2018training}.

Table \ref{table:detailed cifar100 results} presents the detailed experimental results for the CIFAR100 experiments.

\begin{table}
\centering
\begin{tabular}{ccc}
                 & Top1 acc. (\%) & Model size (MB) \\ \hline
\texttt{UDC}              & \textbf{79.24}         & \textbf{0.553}           \\
Gong et al. [27] & 77.84         & 0.57            \\ 
HAQ              & 77.07         & 0.6 \\ \hline \hline
\texttt{UDC}              & \textbf{79.71}         & \textbf{0.705}           \\
Gong et al. [27] & 78.73         & 0.76            \\
McDonnel, [51]   & 76.26         & 1.02           \\
HAQ              & 78.11         & 0.8 \\
FBNet            & 78.64         & 2.8 \\
MBNetV2          & 78.15         & 2.5
\end{tabular}
\caption{Detailed CIFAR100 results comparing compressed model size versus accuracy for \texttt{UDC} against SOTA algorithms. Note that HAQ uses non-uniform quantization, such that models produced by HAQ cannot be deployed on commercial NPUs running integer convolutions.}
\label{table:detailed cifar100 results}
\end{table}

\section{ImageNet experiment settings}
\label{section:ImageNet experiment settings}
We use the same search settings as for the CIFAR100 experiments.
% We run the search for 200 epochs, annealing $\tau$ from $0.66$ to $0.1$ using an exponential schedule. We use SGD for $\bm{\theta}$ with learning rate annealed from $0.1$ to $1e-4$ using a cosine schedule and we use ADAM for $\braces{\pi}$ with a constant learning rate of $1e-3$. We increase $\vartheta$ from $0$ to $0.5$ using a linear schedule and we increase $\zeta^t$ from $0.1$ to $1$ using a linear schedule
To finetune the discovered models, we run stage 1 for 126 epochs using SGD and cosine decay with restarts learning rate schedule, cycling between $0.1$ and $1e-4$ at intervals which double after every cycle and beginning with a cycle of 2 epochs. We run stage 2 for 60 epochs, annealing the learning rate from $0.1$ to $1e-4$ using a cosine schedule and then we run stage 3 for 30 epochs, annealing the learning rate from $1e-4$ to $0$ using a cosine schedule. For the $0.5$ and $1$MB target experiments, we use distillation with a teacher model whose architecture is MobileNetV2. We do not use distillation for the $1.25$MB target experiment. For data augmentation, we use the standard ImageNet data pipeline \cite{dong2019searching}, as well as horizontal flipping and label smoothing with smoothing coefficient $0.1$. We use $\ell_2$ regularization with a coefficient of $1e-4$.

Table \ref{table:detailed imagenet results} shows the detailed ImageNet results.

\begin{table}
\centering
\begin{tabular}{ccc}
                        & Top1 acc. (\%) & Model size (MB) \\ \hline
\texttt{UDC}                     & \textbf{64.13}        & \textbf{0.5}             \\
MCUNet             & 63.5          & 0.67            \\ \hline \hline
\texttt{UDC}                     & \textbf{66.61}        & \textbf{0.9}             \\
MCUNetV2          & 64.9          & 0.99            \\
Choi et al., [18]       & 64.1          & 0.94            \\ 
MCUNetV2                & 64.9          & 0.99 \\ \hline \hline
\texttt{UDC}                     & \textbf{72.05}        & \textbf{1.27}            \\
Choi et al., [18]       & 65.8          & 1.35            \\
MCUNet & 70.7          & 1.57            \\
Gong et al., [27]       & 68.38         & 1.44            \\
Uhlich et al., [60]     & 69.74         & 1.55   \\ \hline \hline
APQ                     & 72.1          & 4.26 \\
FBNet                   & 73.3          & 16.4 \\
FBNetV2                 & 68.3          & 22.89
\end{tabular}
\caption{Detailed ImageNet experimental results, comparing compressed model size versus accuracy for \texttt{UDC} and SOTA algorithms.}
\label{table:detailed imagenet results}
\end{table}

\section{Super resolution experiment settings}
\label{section:Super Resolution experiment settings}
We run the search for 300 epochs, with constant $\tau$ set to $0.1$. We use ADAM for $\bm{\theta}$ with learning rate annealed from $1.e-4$ to $1e-5$ using a cosine schedule and ADAM as well for $\braces{\pi}$ with a constant learning rate of $1e-3$. We keep $\vartheta$ constant to $0.25$ and we increase $\zeta^t$ from $0.1$ to $1$ using a cosine schedule.

Table \ref{table:detailed super resolution results} provides detailed results for the super resolution experiment.

\begin{figure}
        \centering
    \begin{subfigure}[t]{0.33\linewidth}
        \centering
        \includegraphics[width=0.95\linewidth]{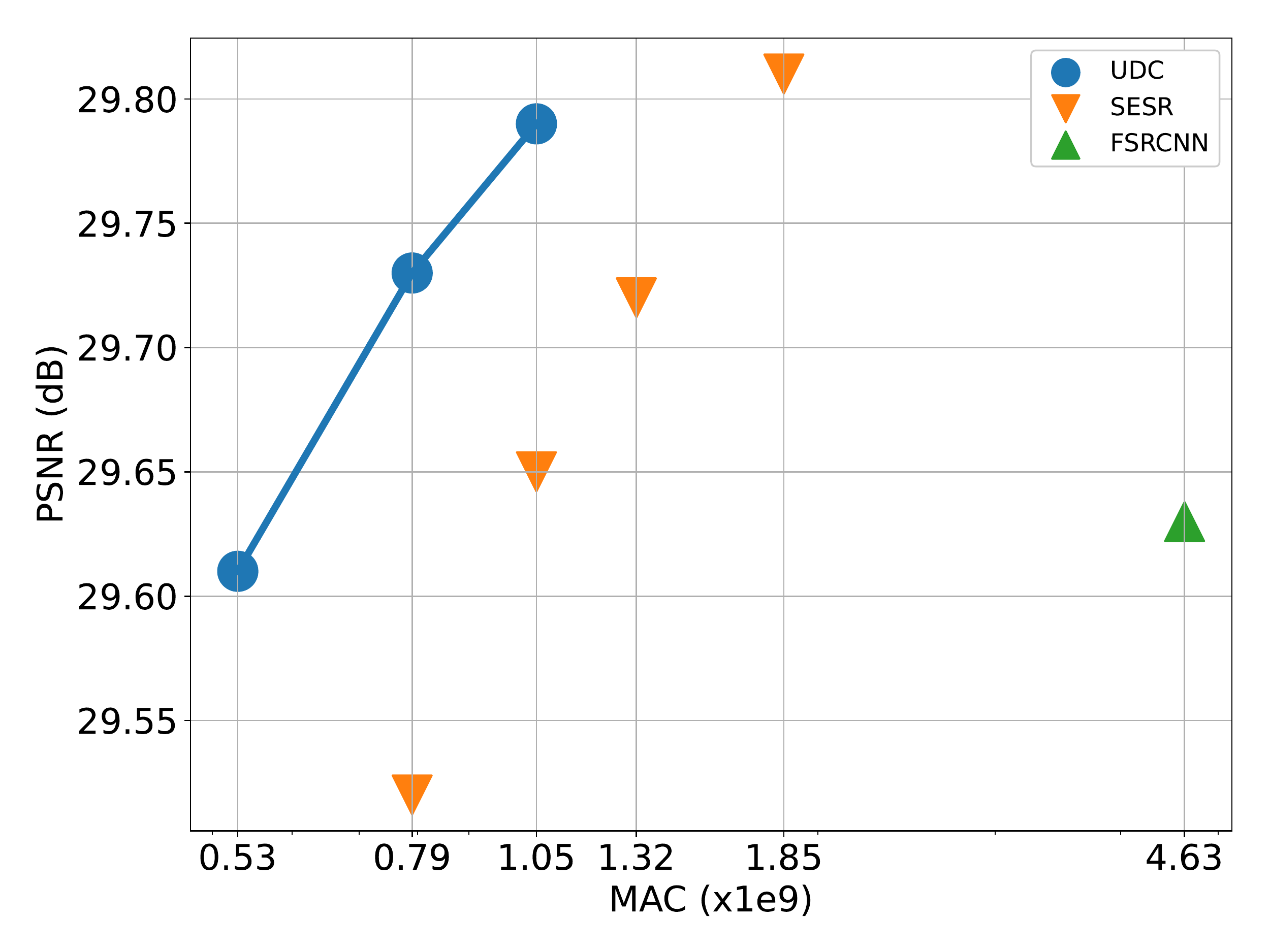}
        \caption{div2k}
        \label{fig:div2k_4x}
    \end{subfigure}%
    ~
    \begin{subfigure}[t]{0.33\linewidth}
        \centering
        \includegraphics[width=0.95\linewidth]{Figures/super_resolution_set14_4x.pdf}
        \caption{set14}
        \label{fig:set14_4x}
    \end{subfigure}%
    ~
    \begin{subfigure}[t]{0.33\linewidth}
        \centering
    \includegraphics[width=0.95\linewidth]{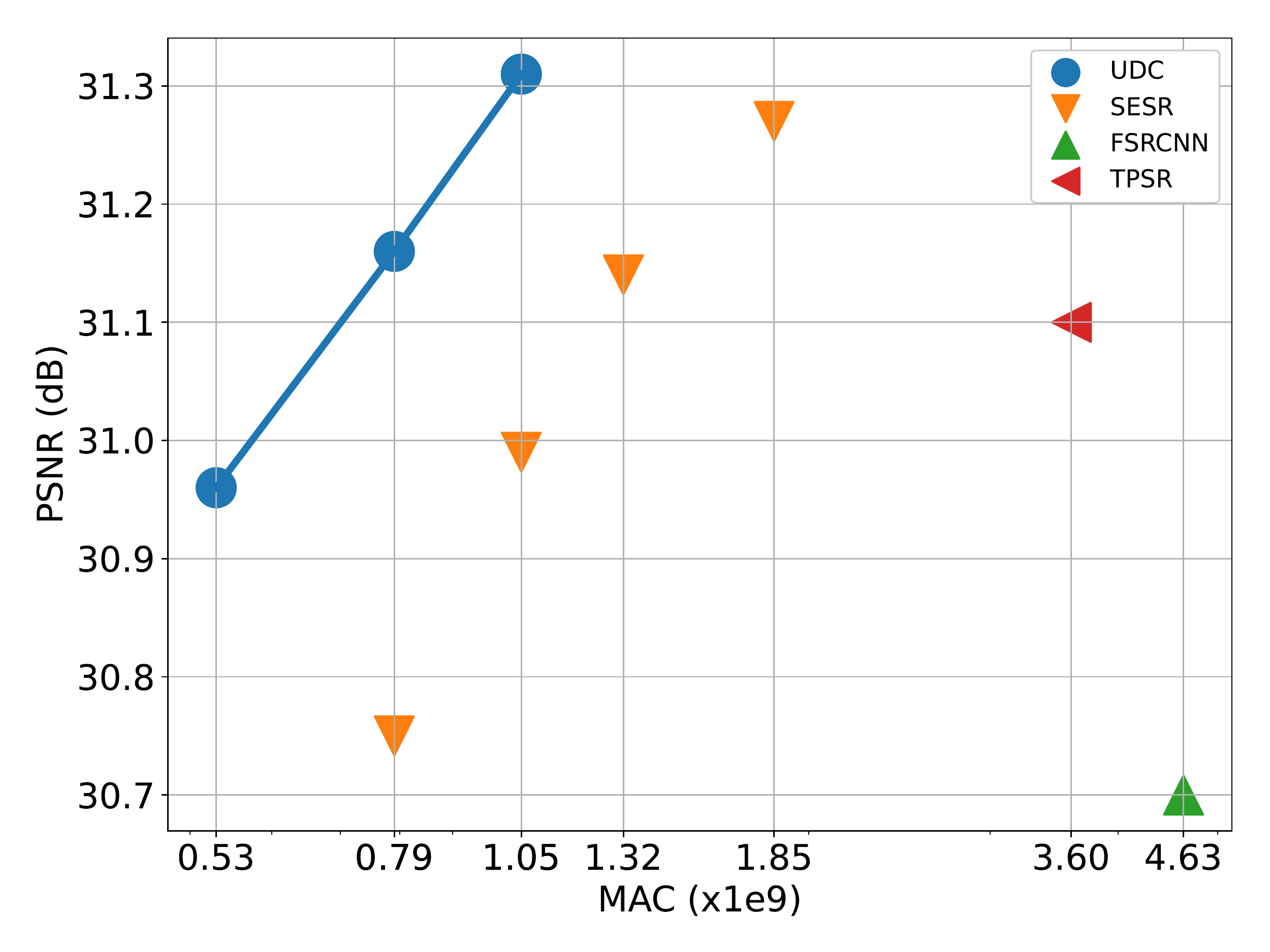}
    \caption{set5}
    \label{fig:set5_4x}
    \end{subfigure}%
\caption{Super resolution results for $4\times$ upsampling. MACs are reported for an input size of $64 \times 64$.}
\label{fig:super resolution results}
\end{figure}

\begin{table}
\centering
\resizebox{\textwidth}{!}{%
\begin{tabular}{ccccccc}
&   \multicolumn{2}{c}{div2k} & \multicolumn{2}{c}{set14} & \multicolumn{2}{c}{set5} \\
& PSNR (dB) & MAC (x1e9) & PSNR (dB) & MAC (x1e9) & PSNR (dB) & MAC (x1e9) \\ \hline
\texttt{UDC} & \textbf{29.61} & \textbf{0.53} & \textbf{27.76} & \textbf{0.53 } & \textbf{30.96} & \textbf{0.53} \\
SESR & 29.52 & 0.79 & 27.69 & 0.79 & 30.75 & 0.79\\ \hline \hline  
\texttt{UDC} & \textbf{29.73} & \textbf{0.79} & \textbf{27.9} & \textbf{0.79} & \textbf{31.16} & \textbf{0.79} \\ 
SESR & 29.52 & \textbf{0.79} & 27.69 & \textbf{0.79} & 30.75 & \textbf{0.79} \\ \hline \hline
\texttt{UDC} & \textbf{29.79} & \textbf{1.05} & \textbf{27.98} & \textbf{1.05} & \textbf{31.31} & \textbf{1.05}\\
SESR & 29.65     & \textbf{1.05 } & 27.81 & \textbf{1.05} & 30.99 & \textbf{1.05} \\ \hline \hline
SESR & 29.72     & \textbf{1.32 } & 27.88 & \textbf{1.32} & 31.14 & \textbf{1.32} \\ 
SESR & \textbf{29.81}     & {1.85 } & \textbf{27.94} & 1.85 & \textbf{31.27} & 1.85\\ 
FSRCNN & 29.63     & {4.63 } & 27.59 & 4.63 & 30.7 & 4.63 \\ 
TPSR & --- & --- & 27.95 & 3.6 & 31.1 & 3.6

\end{tabular}}
\caption{Detailed super resolution experiment results comparing \texttt{UDC} to SOTA efficient super resolution algorithms. MACs are reported for 4x upsampling with an input of size $64 \times 64$.}
\label{table:detailed super resolution results}
\end{table}

% \section{Vela Table}
% \begin{table}
% \centering
% \resizebox{\linewidth}{!}{%
% \begin{tabular}{l|c|c|c|c}
% \toprule
%     & \multicolumn{2}{c|}{Model Size (MB)} & Latency & Acc. \\
%     & Ours & Vela & (ms) & (\%)\\ 
% \midrule
% MBNet V1 1.0 128 8b & $4.3$ & & $7.6$ & $63.3$ \\
% MCUNet~\cite{lin2020mcunet} & $1$ & 0.9 & $4.9$ & $61.8$ \\
% Ours (model size) & $0.5$ & 0.63 & $6.76$ & $63.8$   \\ \hline  
% % Ours (model size \& op count) & $0.49$ & $0.65$ & $4.86$  & $60.1$   \\ \hline
% MBNet V1 0.75 192 8b & $2.6$ & & $5.9$ & $66.1$ \\
% Ours & $1$ & & $7.6$ & $66.7$ \\
% \bottomrule
% \end{tabular}}
% %$^a$ MBNet V1 1.0 128 8b
% \caption{Summary of NNs on Arm Ethos U55-256~\cite{ethos-u55,ethos-vela} NPU}
% %latency for NNs with varying compressibility}
% \label{table:vela-npu}
% \end{table}

\section{Comparison to non-uniform quantization approaches}
\begin{figure}
    \centering
    \includegraphics[width=0.95\linewidth]{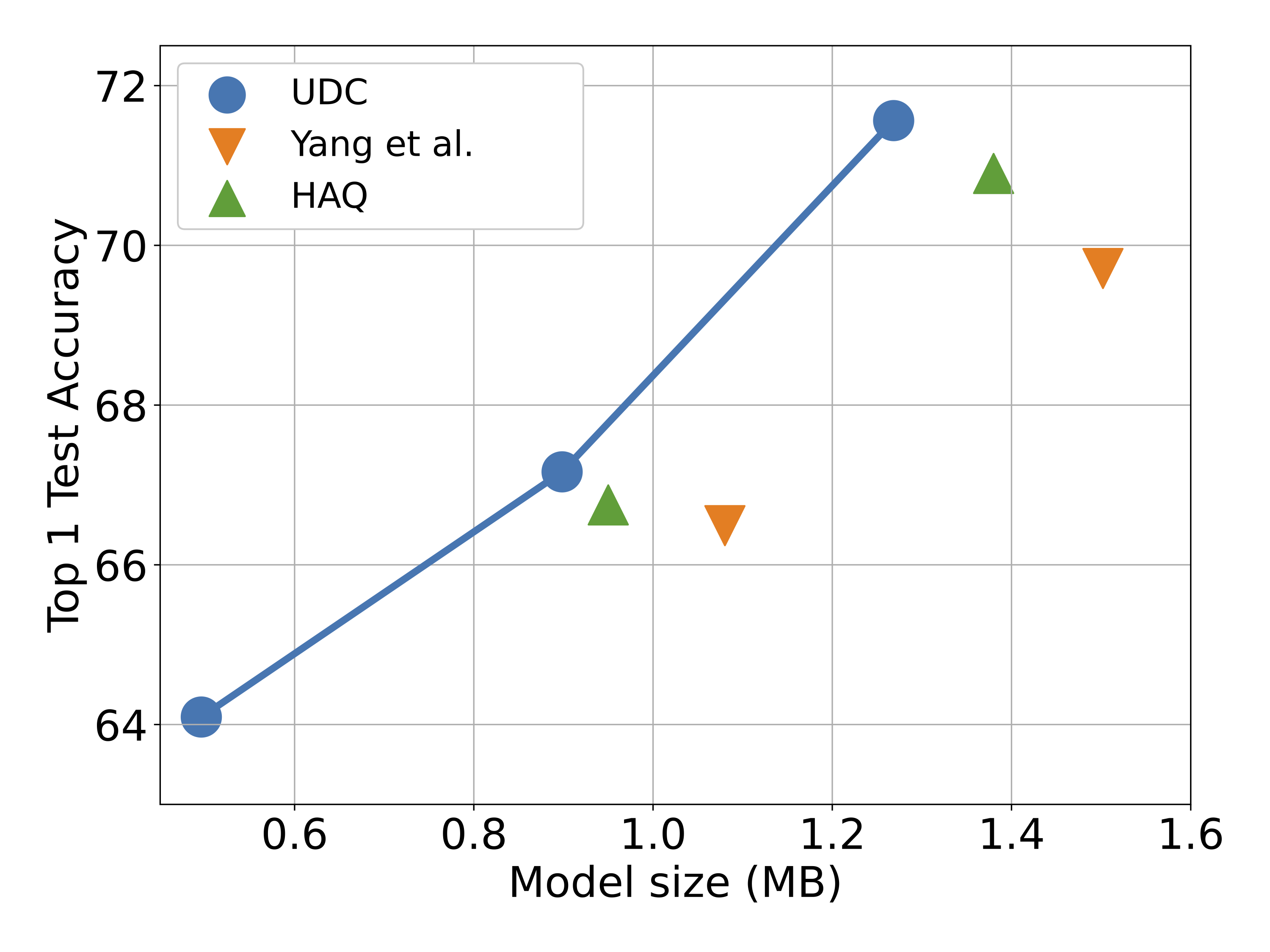}
    \caption{ImageNet test set accuracy vs. compressed model size.}
    \label{fig:non uniform}
\end{figure}

We compare \texttt{UDC} to approaches which employ non-uniform quantization in Fig. \ref{fig:non uniform}. \texttt{UDC} is Pareto-dominant even though it uses uniform quantization and can be deployed on MCUs/NPUs with integer math whereas the other approaches cannot.
\label{section:non uniform quantization}

\section{Comparison to approaches which only do unstructured pruning}
\label{section:comparison with unstructured pruning}
\begin{figure}
    \centering
    \includegraphics[width=0.95\linewidth]{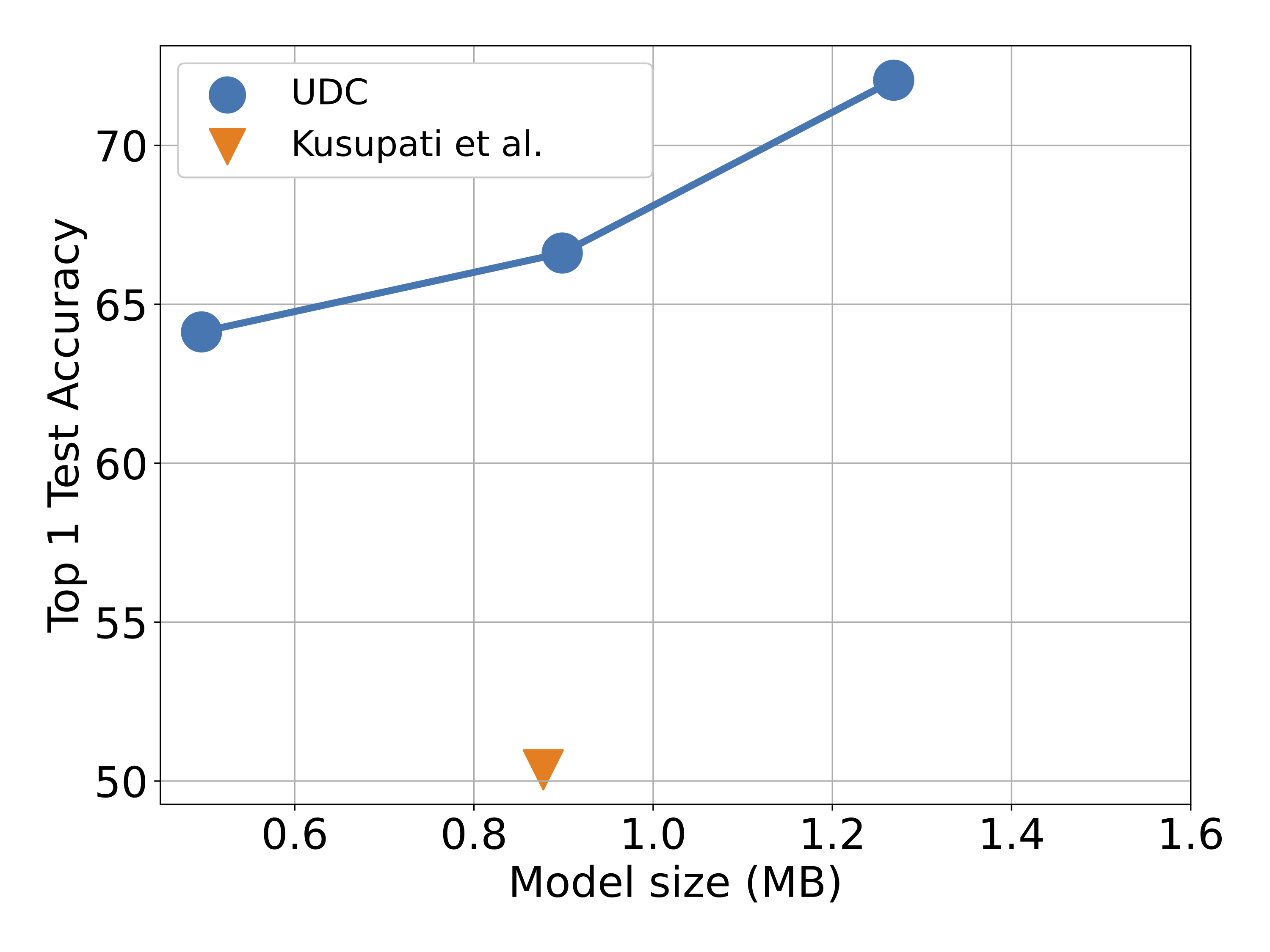}
    \caption{ImageNet test set accuracy vs. compressed model size.}
    \label{fig:unstructured pruning}
\end{figure}

We also compare \texttt{UDC} to a SOTA unstructured pruning algorithm \cite{kusupati2020soft} in Fig. \ref{fig:unstructured pruning}. As the results show, \texttt{UDC} generates much more accurate models.
\label{section:unstructured pruning}

\section{Visualization of design choices}
\begin{figure*}
    \centering
    \begin{subfigure}[t]{0.33\linewidth}
        \centering
        \includegraphics[width=\linewidth]{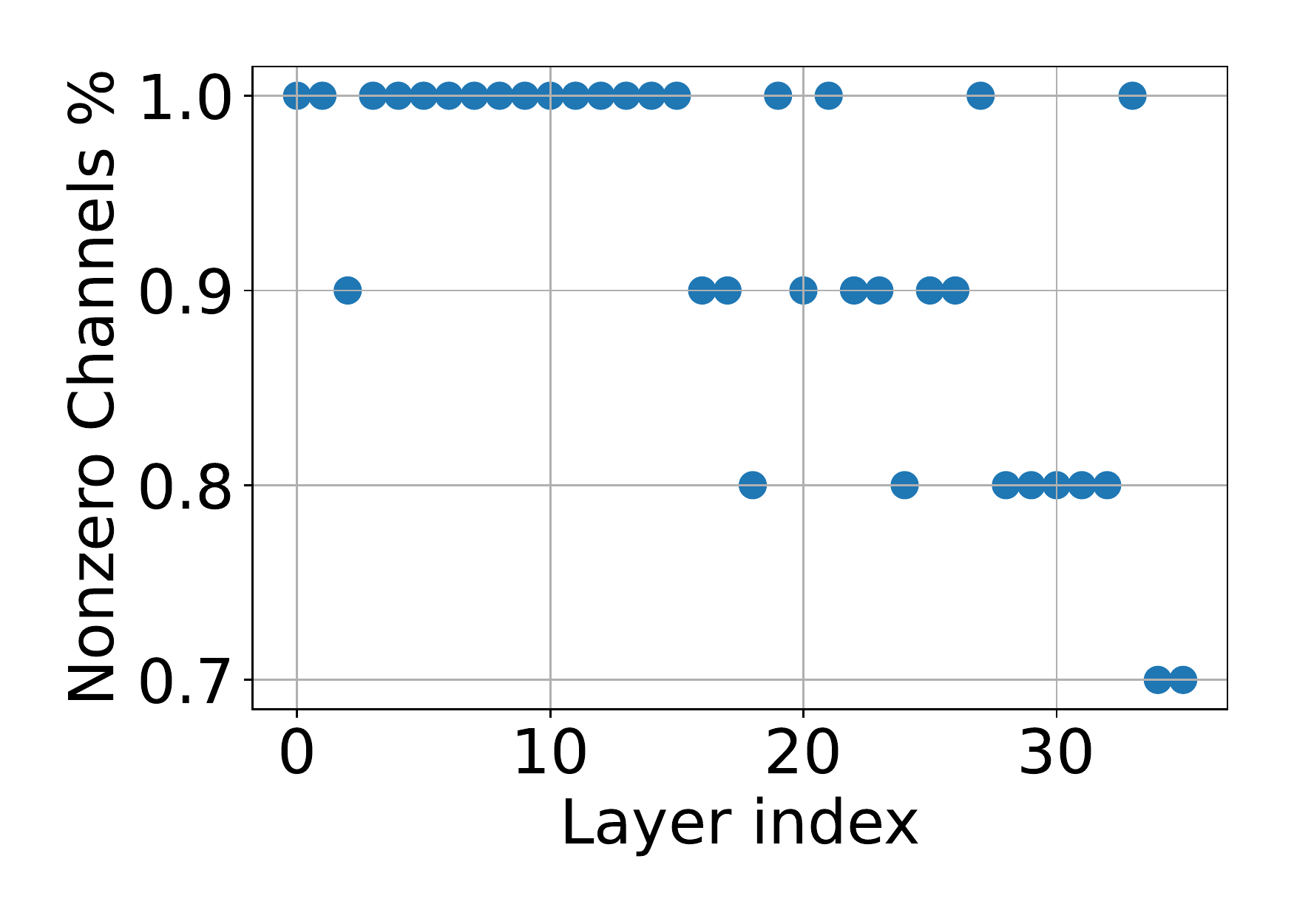}
        \caption{Channel width decisions for 0.5 MB ImageNet experiment}
    \end{subfigure}%
    ~ 
    \begin{subfigure}[t]{0.33\linewidth}
        \centering
        \includegraphics[width=\linewidth]{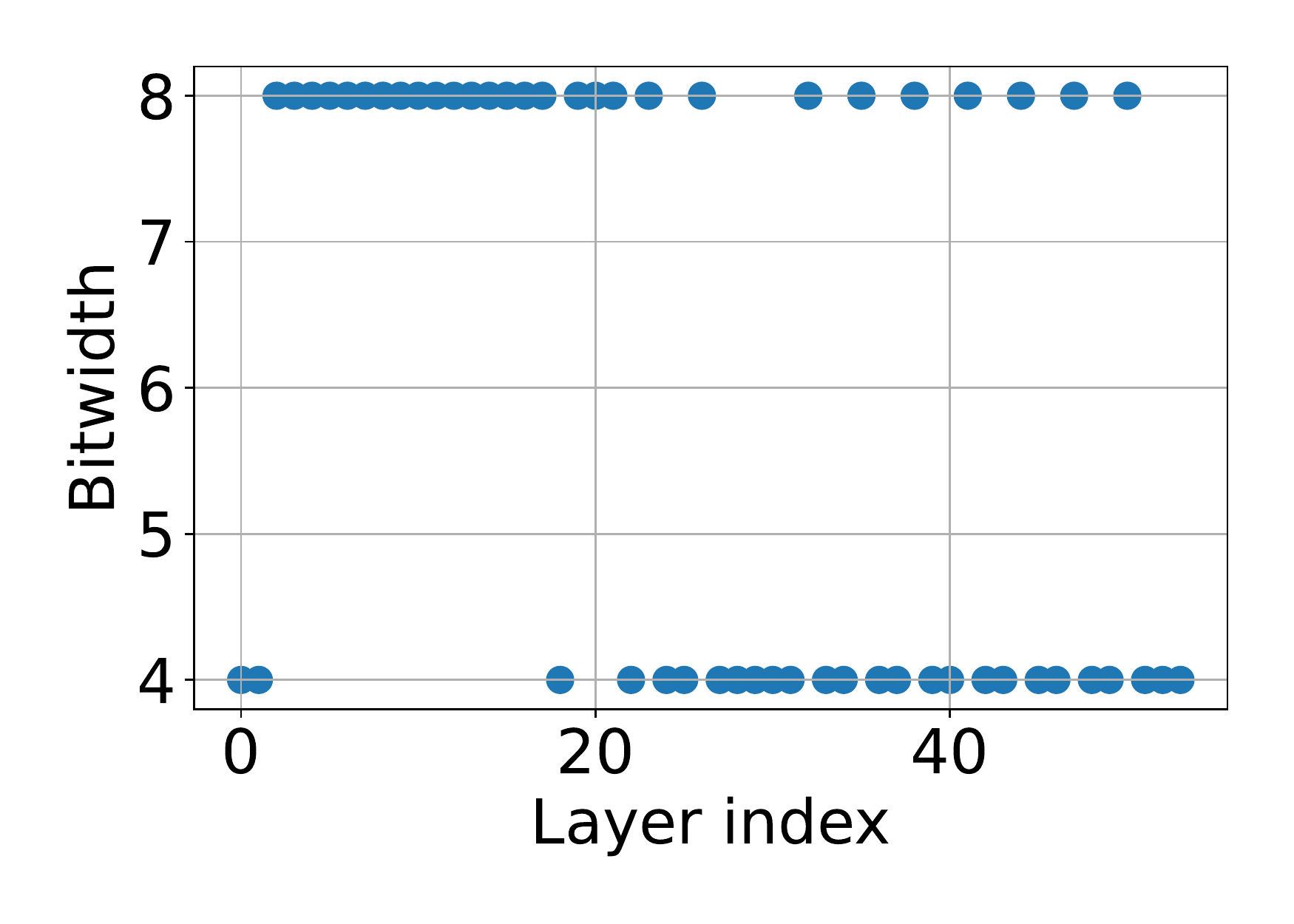}
        \caption{Quantization decisions for 0.5 MB ImageNet experiment}
    \end{subfigure}%
    ~ 
    \begin{subfigure}[t]{0.33\linewidth}
        \centering
        \includegraphics[width=\linewidth]{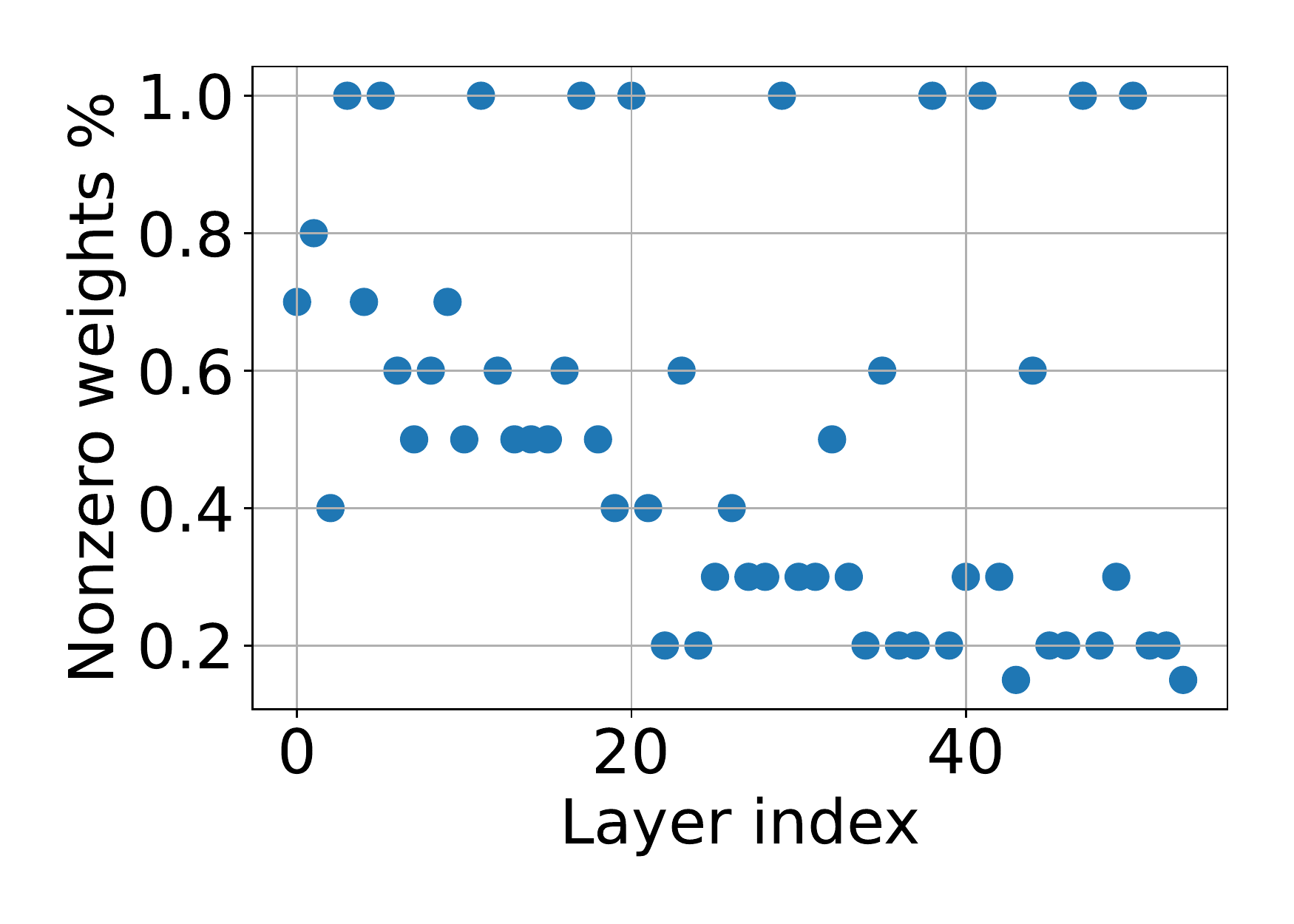}
        \caption{Unstructured pruning for 0.5 MB ImageNet experiment}
    \end{subfigure}
\caption{Model decisions for 0.5MB ImageNet experiment.}
\label{fig:visualization 0.5}
\end{figure*}

\begin{figure*}
    \centering
    \begin{subfigure}[t]{0.33\linewidth}
        \centering
        \includegraphics[width=\linewidth]{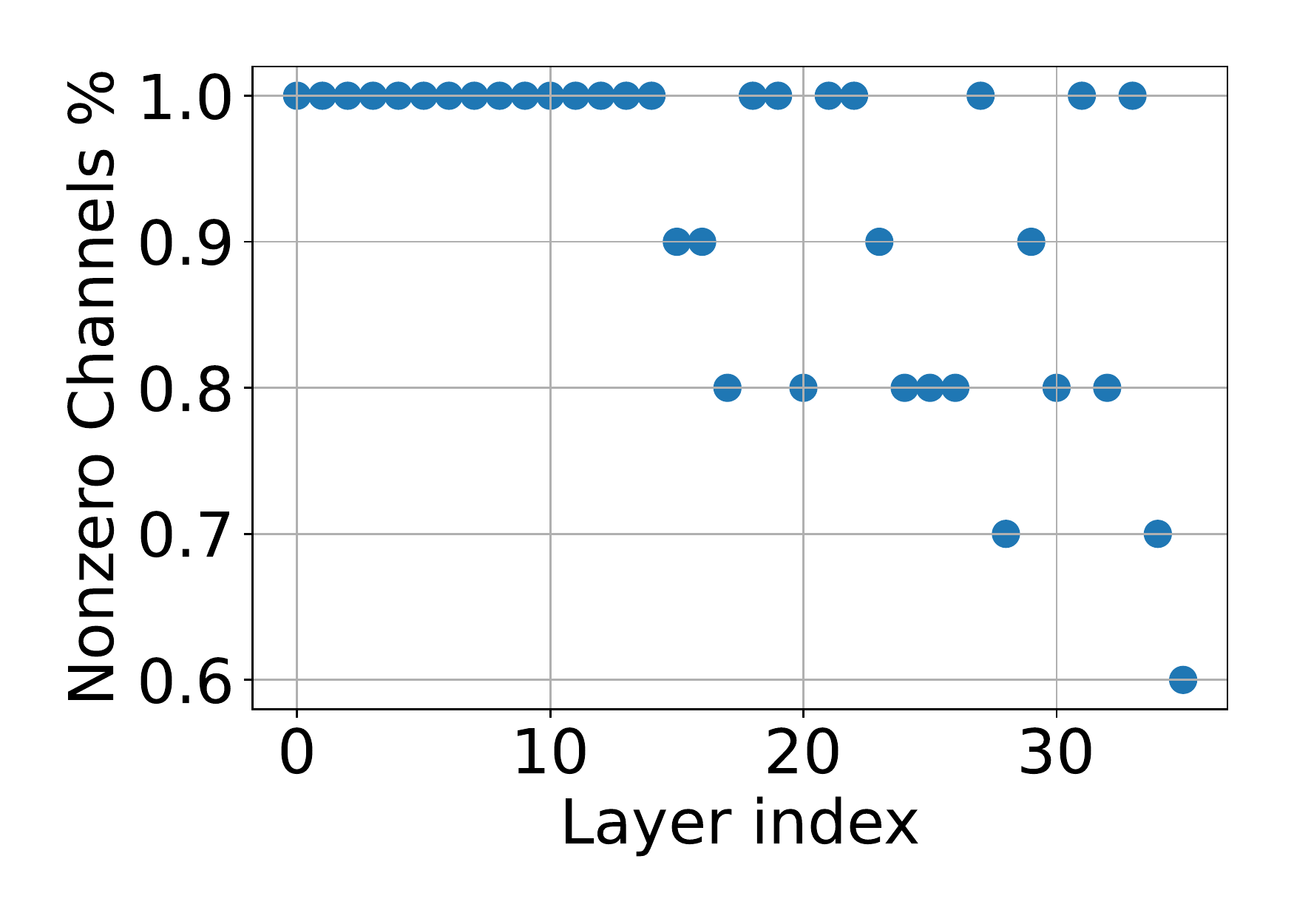}
        \caption{Channel width decisions for 1 MB ImageNet experiment}
    \end{subfigure}%
    ~ 
    \begin{subfigure}[t]{0.33\linewidth}
        \centering
        \includegraphics[width=\linewidth]{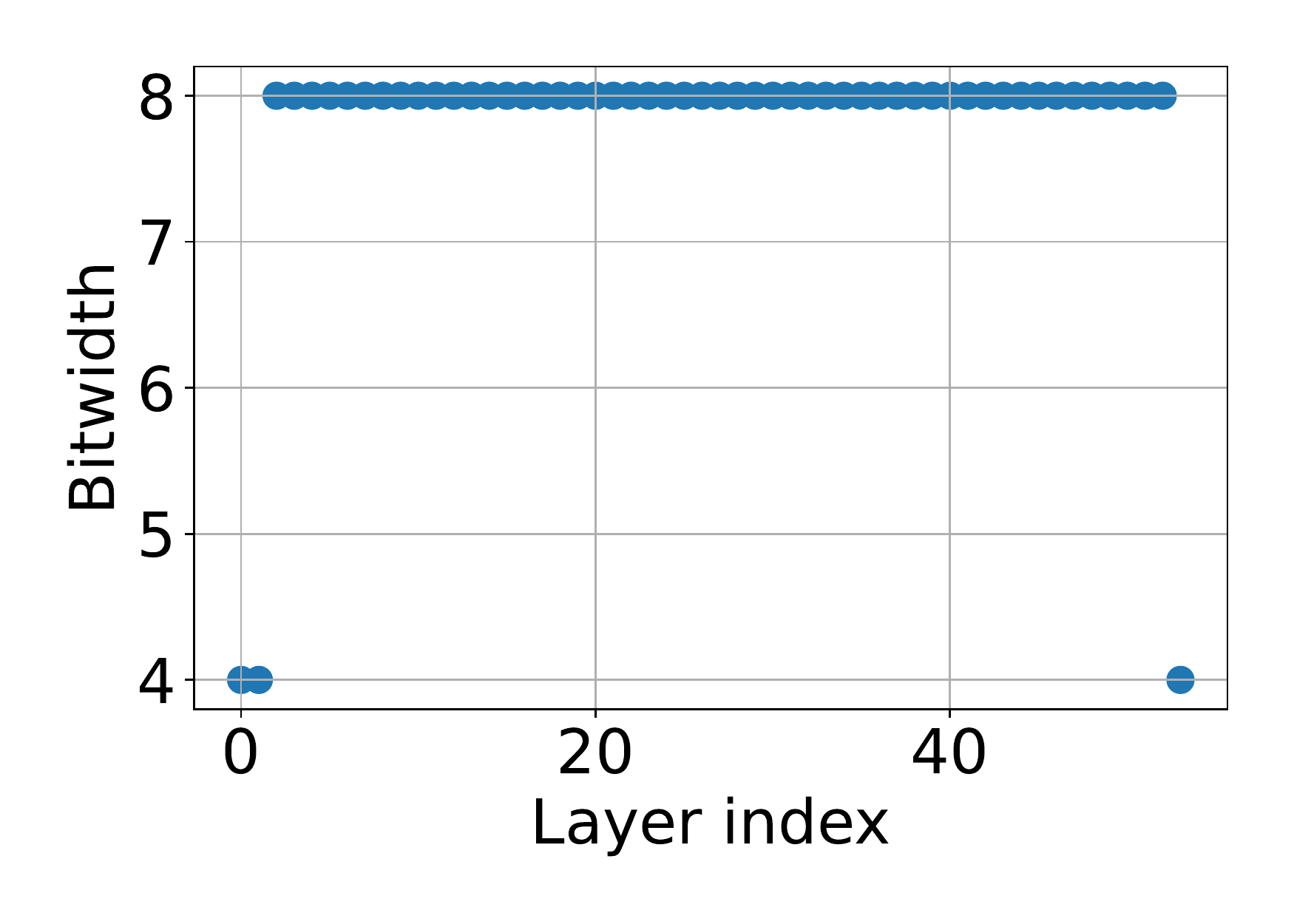}
        \caption{Quantization decisions for 1 MB ImageNet experiment}
    \end{subfigure}%
    ~ 
    \begin{subfigure}[t]{0.33\linewidth}
        \centering
        \includegraphics[width=\linewidth]{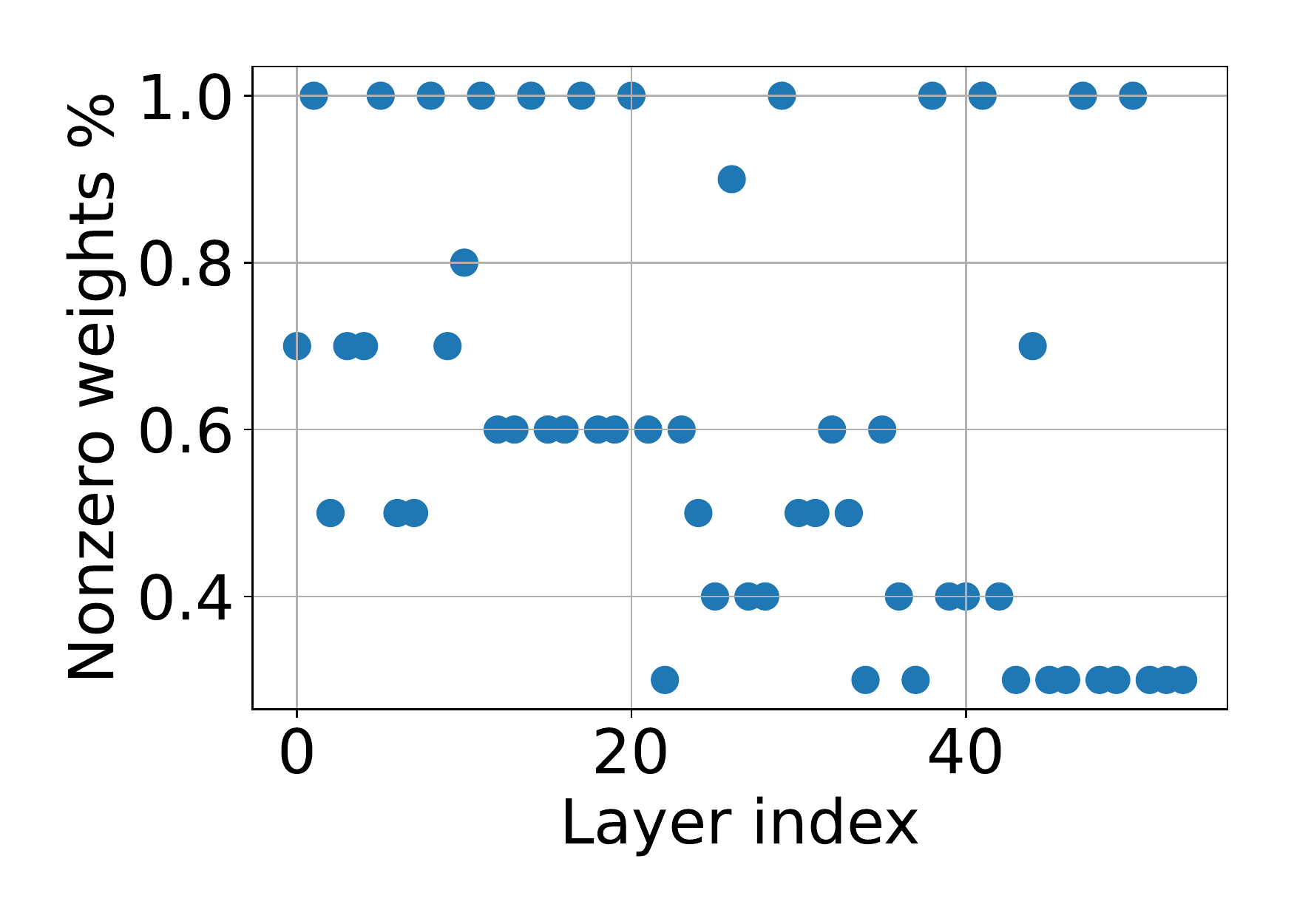}
        \caption{Unstructured pruning decisions for 1 MB ImageNet experiment}
    \end{subfigure}
\caption{Model decisions for 1 MB ImageNet experiment.}
\label{fig:visualization 1}
\end{figure*}

\begin{figure*}
    \centering
    \begin{subfigure}[t]{0.33\linewidth}
        \centering
        \includegraphics[width=\linewidth]{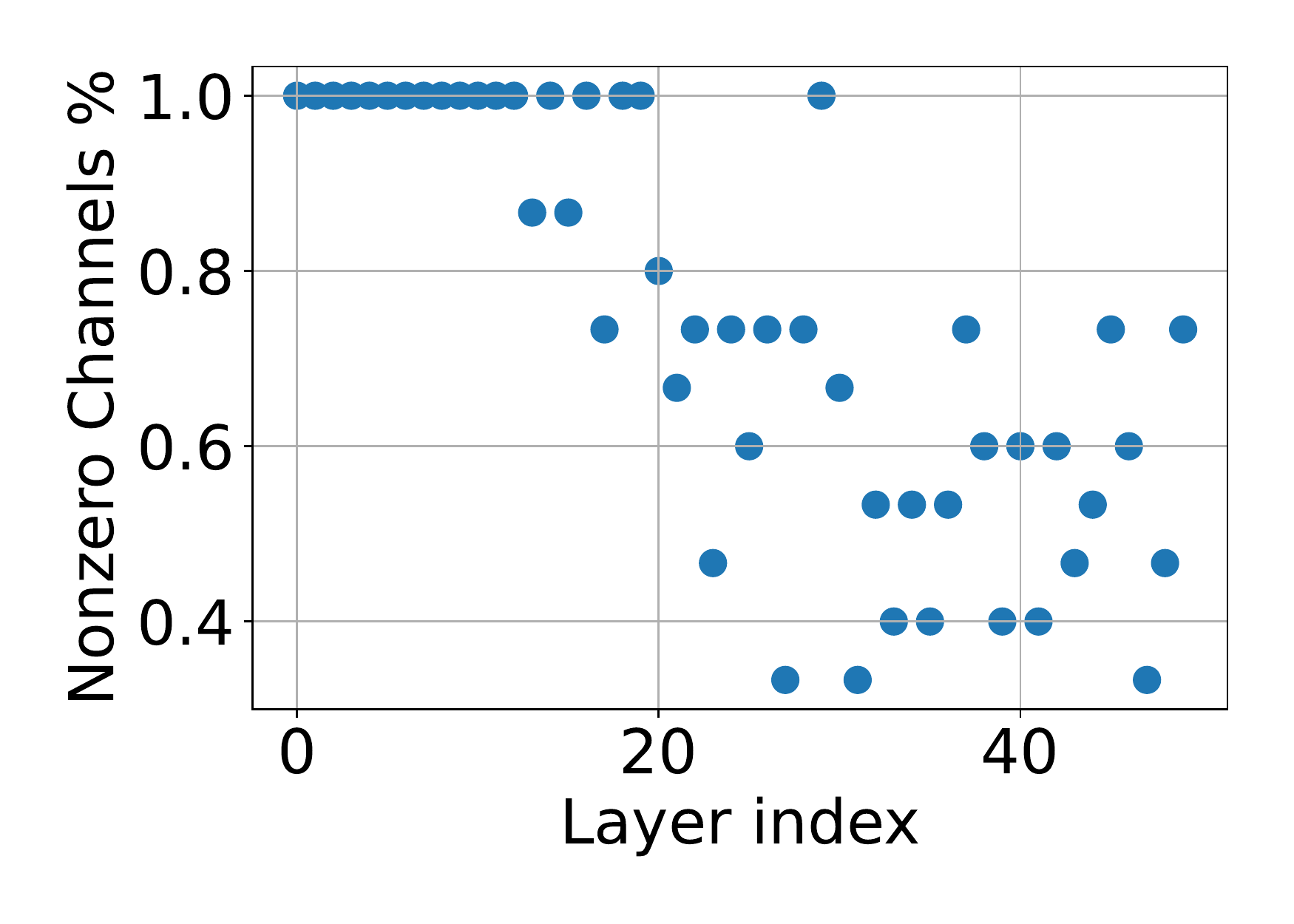}
        \caption{Channel width decisions for 1.25 MB ImageNet experiment}
    \end{subfigure}%
    ~ 
    \begin{subfigure}[t]{0.33\linewidth}
        \centering
        \includegraphics[width=\linewidth]{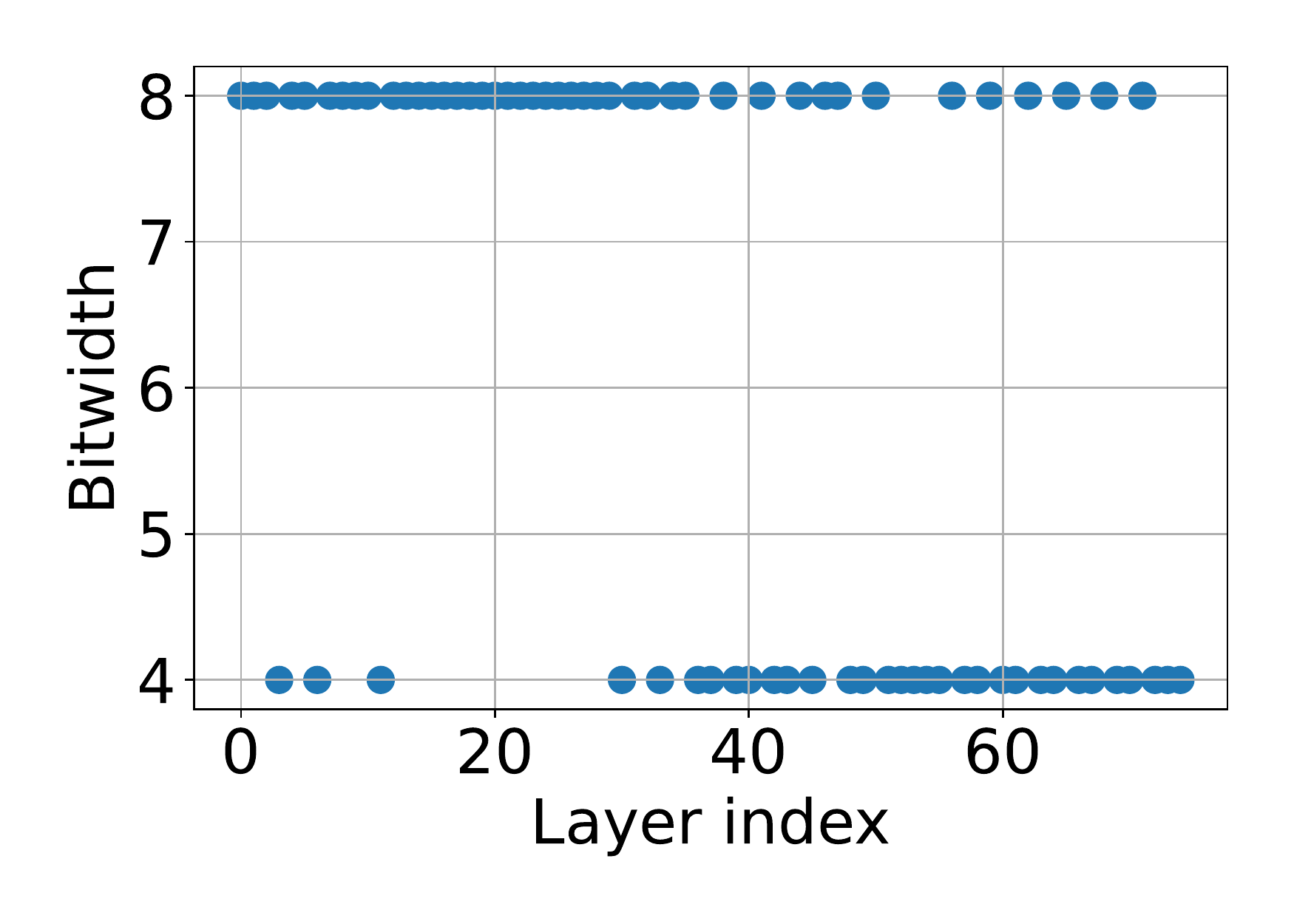}
        \caption{Quantization decisions for 1.25 MB ImageNet experiment}
    \end{subfigure}%
    ~ 
    \begin{subfigure}[t]{0.33\linewidth}
        \centering
        \includegraphics[width=\linewidth]{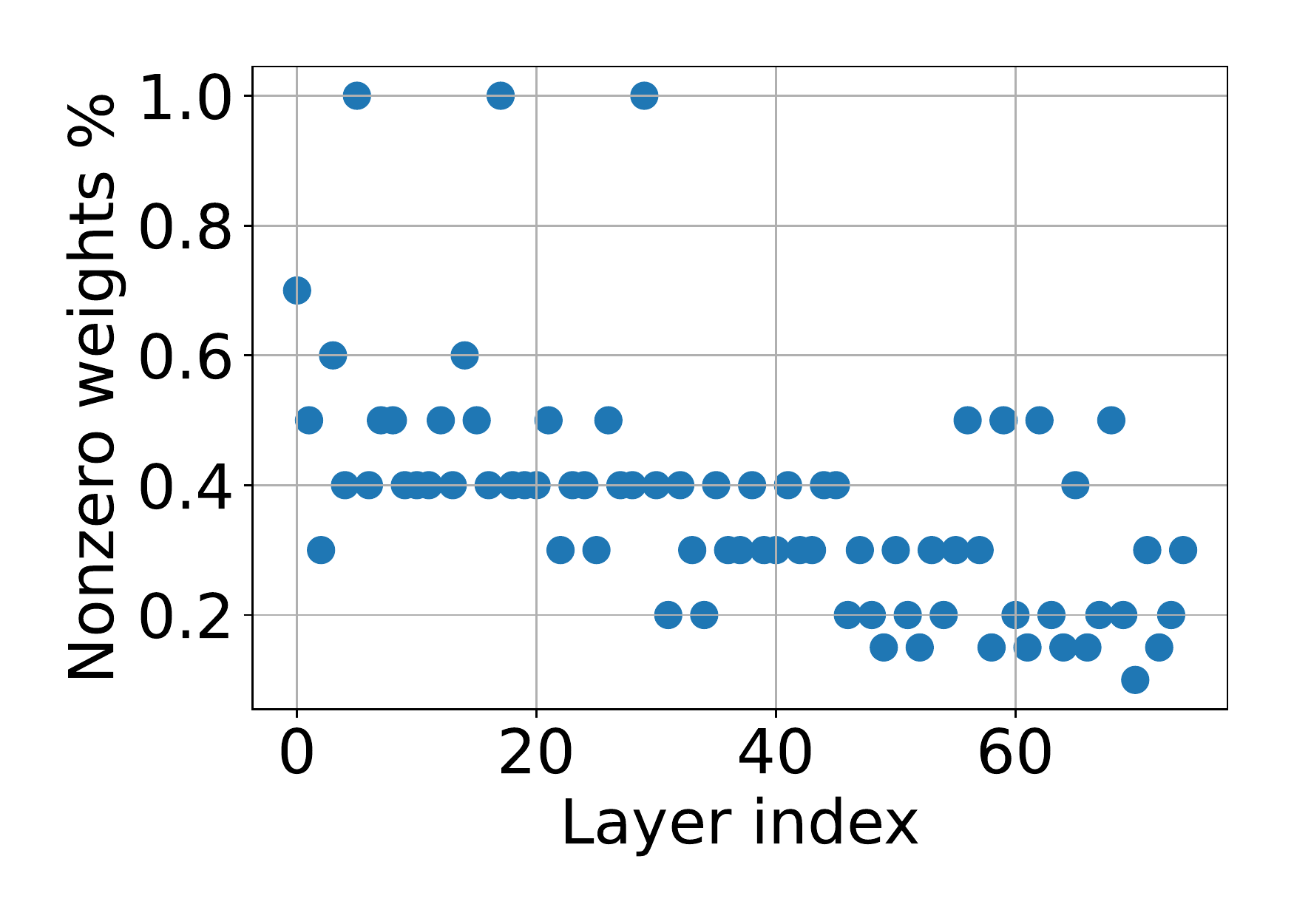}
        \caption{Unstructured pruning decisions for 1.25 MB ImageNet experiment}
    \end{subfigure}
\caption{Model decisions for 1.25 MB ImageNet experiment.}
\label{fig:visualization 1.25}
\end{figure*}

We provide a visualization of the design choices made by \texttt{UDC} for the ImageNet experiments in Fig. \ref{fig:visualization 0.5}-\ref{fig:visualization 1.25}.

% \section{Deployment of models on NPUs}
% While NPUs only support integer math convolutions, some element-wise floating point operations are supported. 

\end{document}